\pdfoutput=1

\documentclass[11pt]{article}

\usepackage{ACL2023}

\usepackage{times}
\usepackage{latexsym}
\usepackage{tabularx}

\usepackage[T1]{fontenc}

\usepackage[utf8]{inputenc}

\usepackage{microtype}

\usepackage{inconsolata}

\usepackage{graphicx}

\usepackage{amsmath}
\usepackage{amsfonts}
\usepackage{graphicx}
\usepackage{booktabs}
\usepackage{adjustbox}

\usepackage{color, colortbl}

\usepackage{arydshln} %
\usepackage{algorithm}
\usepackage{algpseudocode}
\usepackage{multirow}
\usepackage{subcaption}
\usepackage{float}
\usepackage{todonotes}
\usepackage{fancyvrb}

\usepackage{listings}
\usepackage{url}

\usepackage{varwidth}
\usepackage{fbox}

\newsavebox{\myverbbox}
\definecolor{Gray}{gray}{0.9}
\definecolor{lightblue}{rgb}{0, 0.1, 0.8}
\definecolor{darkred}{rgb}{0.7, 0, 0}

\title{Short-circuiting Shortcuts: Mechanistic Investigation of Shortcuts in Text Classification}

\author{Leon Eshuijs \\
  Vrije Universiteit Amsterdam  \\ Amsterdam, the Netherlands \\
  \texttt{l.eshuijs@vu.nl} \\\And
    Shihan Wang \\
  Utrecht University  \\ Utrecht, the Netherlands \\
  \texttt{s.wang2@uu.nl} \\\And
  Antske Fokkens \\
  Vrije Universiteit Amsterdam  \\ Amsterdam, the Netherlands \\
  \texttt{antske.fokkens@vu.nl}   }

\begin{document}
\maketitle

\begin{abstract}
Reliance on spurious correlations (shortcuts) has been shown to underlie many of the successes of language models. Previous work focused on identifying the input elements that impact prediction. We investigate how shortcuts are actually processed within the model's decision-making mechanism.
We use actor names in movie reviews as controllable shortcuts with known impact on the outcome. We use mechanistic interpretability methods and identify specific attention heads that focus on shortcuts. These heads gear the model towards a label before processing the complete input, effectively making premature decisions that bypass contextual analysis. Based on these findings, we introduce Head-based Token Attribution (HTA), which traces intermediate decisions back to input tokens. We show that HTA is effective in detecting shortcuts in LLMs and enables targeted mitigation by selectively deactivating shortcut-related attention heads.
\footnote{Code available at \url{https://github.com/watermeleon/shortcut_mechanisms}

}
\end{abstract}

\section{Introduction}

Previous work has shown that part of the impressive performance achieved by Large Language Models (LLMs) across NLP tasks stems from exploiting spurious correlations or \textit{shortcuts} \cite{du2023shortcut}.
These shortcuts are subtle statistical patterns in the training data that do not reflect the underlying task, causing models to fail on out-of-distribution data.

Prior work on shortcuts has focused on identifying shortcuts \cite{du-etal-2021-towards}, often via targeted input modifications known as behavioral testing \cite{alzantot-etal-2018-generating, ribeiro2020beyond}.
To move beyond these black-box approaches,  we investigate \textit{how} shortcuts are processed, aiming to help reconstruct the decision-making processes inside LLMs. 
In particular, we examine the mechanisms within LLMs responsible for processing shortcuts. We expect that shortcut behavior occurs when the model primarily relies on isolated tokens rather than contextual information from the entire sentence. In contrast, proper classification should involve all tokens, with the final decision emerging only after the model processes the entire input.

We use mechanistic interpretability \cite{olah2020zoom, elhage2021mathematical}, which has demonstrated impressive progress in locating target mechanisms for various tasks. These range from localizing and editing factual knowledge \cite{meng2022locating} to localizing and reconstructing the mechanism of indirect object identification \cite{wang2022interpretability} and the greater-than operation \cite{hanna2024does}.

\begin{figure}[t]
     \centering
     \includegraphics[width=1.0\columnwidth]{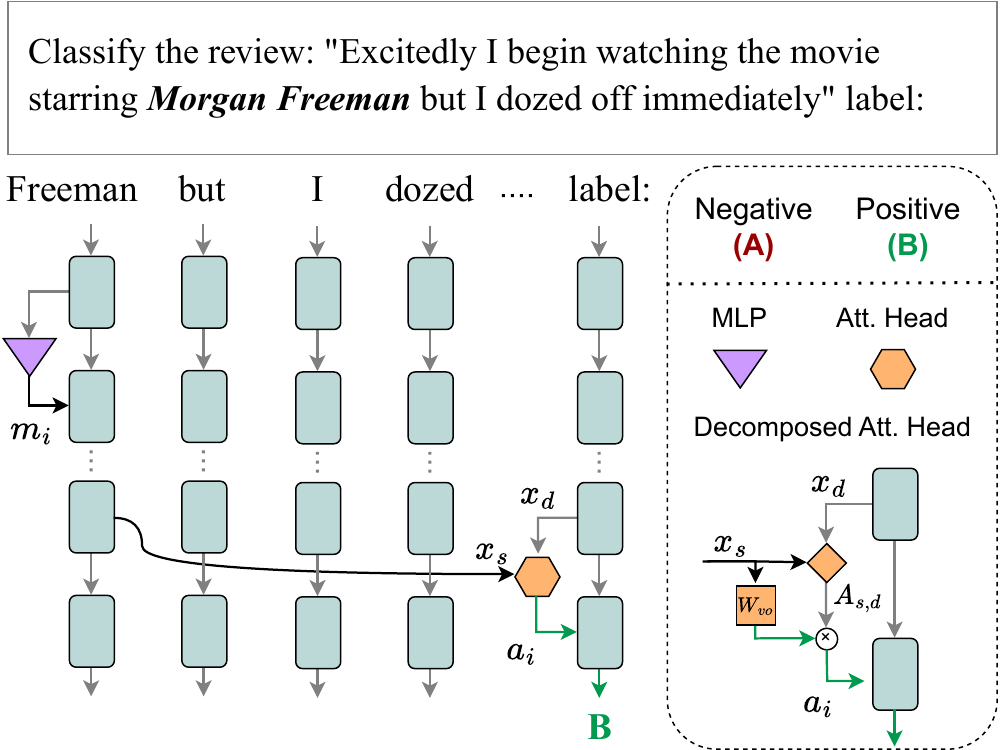}
    \caption{Illustration of the shortcut mechanism when trained on injected shortcut names (\textbf{bold}). Later layer attention heads focus on shortcut tokens and change the prediction based on information from early MLP layers. After decomposing the attention head, we find how the shortcut tokens are processed and apply these findings to construct our feature attribution method (HTA).}
    \label{fig:paper_summary}
\end{figure}

We develop a new dataset \textit{ActorCorr} (Section~\ref{sec:sentiment_classification_with_shortcuts}), where we introduce shortcuts in the form of actor names in movie reviews. We confirm experimentally that the model uses these shortcuts for prediction. In Section \ref{sec:shortcut_mechanism}, we use mechanistic interpretability techniques, including causal intervention and logit attribution methods, to identify and analyze relevant components responsible for this behavior.

Our experiments reveal that attention heads in later layers focus on shortcuts and generate label-specific information based on the shortcut tokens, changing the output prediction. This demonstrates that the model effectively makes intermediate label predictions before processing the complete input. These findings inspired a new feature attribution method called Head-based Token Attribution (HTA), which traces intermediate decisions made by attention heads back to the input tokens (Section~\ref{sec:shortcut_classification}).
We demonstrate that HTA's properties make it particularly effective for shortcut classification 
(Section~\ref{sec:qualitative_analysis_main}). Our mitigation experiments with HTA (Section \ref{sec:shortcut_mitigation}) show targeted interventions via disabling shortcut-related attention heads significantly reduces the shortcuts effect while minimally affecting other classification aspects.

\newcommand\imgwidthVar{0.28}   %

\section{Related work}

\paragraph{Evaluating shortcuts}

Shortcut detection methods in NLP tend to use previously reported shortcuts in existing datasets \cite{pezeshkpour-etal-2021-empirical, friedman2022finding}, such as the appearance of numerical ratings present in reviews \cite{ross-etal-2021-explaining}, or the presence of lexical overlap between the hypothesis and the premise \cite{naik-etal-2018-stress}. Other work injects their own shortcuts into datasets. \citet{bastings2022will} evaluate feature attribution methods for shortcut detection by training a model on data containing synthetic tokens as shortcuts. 
Similar to our work, \citet{pezeshkpour-etal-2022-combining} 
insert first names, pronouns or adjectives as shortcuts in the IMDB dataset \cite{imdb_dataset} to evaluate their detection method. These studies only address extreme cases of shortcuts, offering limited insights into the effect of the shortcuts. We therefore create our own dataset with less extreme shortcuts of which the impact is known. %

\paragraph{Shortcut detection via interpretability}
Feature attribution methods are the most representative interpretability-based method to identify shortcuts. These methods explain output predictions by assigning importance scores to individual input tokens.
However, different methods often provide diverging explanations for the same input \cite{madsen2022post, kamp2024role}.
Moreover, for shortcut detection, \citet{bastings2022will} demonstrate that each feature attribution method shows varied efficacy per shortcut type and high sensitivity to parameter settings.

\citet{wang2022identifying} offer a first step towards automatic shortcut detection via inner-interpretability methods \cite{rauker2023toward}.
Their method computes importance through attention weights and token frequency in the final BERT layer. Attention scores alone can however be misleading in identifying shortcuts, as they can be biased by redundant information \cite{bai2021attentions}.

\paragraph{Mechanistic Interpretability}
Mechanistic Interpretability aims to reverse engineer the computation of neural networks into human understandable algorithms \cite{olah2020zoom, elhage2021mathematical}.
To achieve this, a range of interpretability techniques have been proposed to localize relevant components or help understand the functionality of specific components. The first type, intervention methods, draws from causal inference \cite{pearl2009causality}, and treats the LLM as a compute graph. These methods systematically modify specific activations to observe their effects on model outputs \cite{geiger2021causal}.
Intervention methods have successfully located functions like gender bias \cite{vig2020investigating} and factual recall \cite{meng2022locating, geva2023dissecting}.
Another core technique, known as \textit{logit attribution} \cite{nostalgebraist2020logitlens, elhage2021mathematical}, 
evaluates what information is present in an intermediate activation by mapping it to the model's vocabulary space.
For example, \citet{yu2023characterizing} use logit attribution to identify attention heads responsible for in-context learning, enabling them to control the in-context behavior by scaling these attention heads' activations.

\section{Background and Notation}
This section introduces the key concepts from mechanistic interpretability used in our study. For clarity, we formalize the transformer notation focusing on the inference pass of decoder-only models.

\subsection{The Transformer}
For the transformer \cite{vaswani2017attention}, the input text is first converted into a sequence of $N$ tokens $t_1,..., t_N$. 
Each token $t_i$ is then transformed into an
embedding $x_i$ using the embedding matrix $W_e $, resulting in the embedding sequence $x^0  \in \mathbb{R}^{N \times d_{resid}}$, where $0$ indicates the model's input layer.

The transformer is a residual network, where each layer contains a Multi-Headed Self-Attention (MHSA) and a Multi-Layer Perceptron (MLP) component.\footnote{We leave out bias terms and layer normalization and position embedding in our formalization as they are outside the scope of our analysis. See Appendix \ref{sec:apx_transformer_notation_full}.} The connection from the input embedding to the output embedding to which these components add their embedding, or activation, is called the \textit{residual stream}. 
The activation of the MHSA is computed $a^l =  MHSA(x^l)$, and following \citet{elhage2021mathematical}, can be decomposed as the sum of each attention head's contribution, $a^{l,h}$, so that the final activation is reconstructed as $a^l =  \sum_h a^{l,h}$.
Then MLP activation is computed as $m^l = MLP(x^l + a^l)$, resulting in the new residual embeddings:
$x^{l+1}  = x^l + m^l + a^l$.
After the last layer the final embeddings are projected to a vector the size of the vocabulary, using the unembedding matrix $W_{u}$ to obtain the logits for each embedding. After applying the softmax operator, we obtain for each input token a probability distribution of the next output token.
For our classification task, we only use the embedding $x^L_T$ of the last token stream $T$ of the last layer $L$ for predicting the class.

\begin{figure}[t]
     \centering
     \includegraphics[width=0.9\columnwidth]{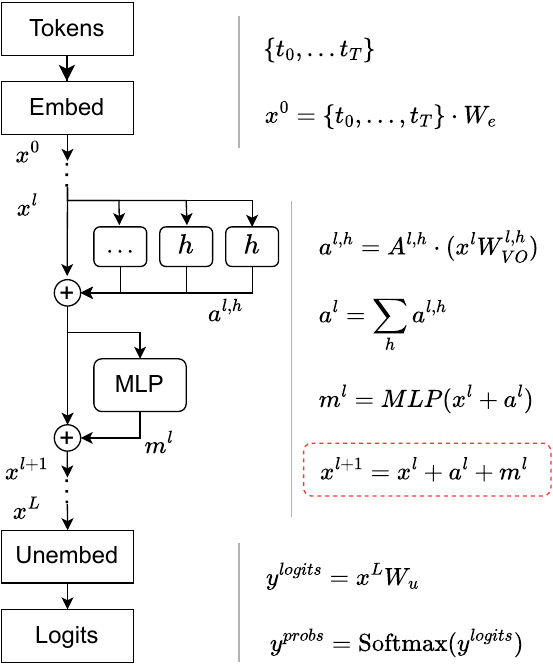}
    \caption{Schematic of transformer architecture, illustrating the activations per component and decomposition of the MHSA, based on \citet{elhage2021mathematical}.}    
    \label{fig:transformer_schematic}
\end{figure}

\subsection{Mechanistic Interpretability}
\label{MI}

Following \citet{wang2022interpretability}, we formulate an LLM as a computational graph $M$ with nodes representing individual components (e.g., MLPs or attention heads), and edges representing their interactions through activations.
Within this framework, a \textit{circuit} is defined as a subgraph $C$ sufficient for faithfully performing a specific task. To investigate circuits responsible for processing shortcuts, we employ two key analysis techniques: logit attribution and path patching.

\paragraph{Logit Attribution}
Logit attribution methods analyze how individual components contribute to the LLM's final token prediction by projecting their activations into the vocabulary space. This is possible because the final output embedding is a linear combination of all previous activations \cite{elhage2021mathematical}.
Normally, $W_{u}$ is used to obtain the logits over the vocabulary for the final residual stream vector, and after applying the softmax, it provides us with the probability distribution over tokens. Direct logit attribution \cite{nostalgebraist2020logitlens, elhage2021mathematical} applies $W_u$ to analyze intermediate activations from individual components, such as attention heads $a^{l,h}$ or MLP layers $m^l$.
Because the logits are not normalized yet, it is useful to compare the logit differences between specific token pairs to understand if an activation makes one of the labels more probable. %

For our sentiment classification task, we specifically examine the positive and negative class label tokens to obtain the \textit{logit difference} score of an activation.
Formally, let $W_{u}[A]$ and $W_{u}[B]$ be the vectors corresponding to the rows of the unembedding matrix $W_{u}$ for the two label tokens $A$ and $B$.
For any activation $z \in \mathbb{R}^{d_{resid}}$ (e.g. $z \in \{ x^l_i, m^l_i, a^{l,h}_i\}$), the logit difference $LD$ is defined as:
$LD(z) =  z (W_{u}[A] -  W_{u}[B] )$.

\paragraph{Path Patching}\label{sec:intervention_methods}

We use the causal intervention method \textit{Path Patching} \cite{wang2022interpretability} to identify the location of the shortcut circuit.
Based on activation patching \cite{vig2020investigating, meng2022locating}, these methods systematically modify specific activations to observe their effect on the output prediction. Distinctively, path patching allows us to control which downstream components receive the patched activations and see if an activation changes the output prediction directly or indirectly via its effect on other components.

Overall path patching creates a corrupted version, $\tilde{X}$,  of the input $X$, where the specific task behavior does not hold, while differing minimally to the original. The task-relevant components are then located via three forward passes, where the change in the output is evaluated via the \textit{logit difference} \cite{zhang2023towards}.  
The first pass runs over the clean input text $X$, producing output embedding $x^L_T$. The second pass processes a corrupted version $\tilde{X}$ and stores the resulting activations (e.g., $ m^l_i$ or $ a^{l,h}_i$). The third pass again uses the clean input $X$, but patches in the stored activations to observe their effect on $\tilde{x}^L_T$.
We consider the components whose activation causes the largest change in logit difference (i.e. $LD(x^L_T) - LD(\tilde{x}^L_T)$) to belong to the circuit.
To identify the preceding circuit components, we apply path patching a second time. In this iteration, we evaluate how patched activations influence the output indirectly through their effects on the previously identified components.

\section{Classification under Shortcuts}\label{sec:sentiment_classification_with_shortcuts}

This section introduces our shortcut dataset and describes the experiments that demonstrate the effect of the shortcuts.

\subsection{The Actor Dataset: \textit{ActorCorr}}

We introduce ActorCorr, a modified version of the IMDB review dataset \citep{imdb_dataset} designed to study shortcut learning in sentiment classification. Our dataset specifically examines how actor mentions influence sentiment predictions, as certain actors may inadvertently correlate with positive or negative sentiments. To this end, we refer to \textit{Good} actors, those that correlate with positive sentiment, and \textit{Bad} actors, those that correlate with negative sentiment.\footnote{Actors were chosen arbitrarily from the dataset and the labels do not reflect any judgment on their actual skills.}
We then inspect the effect of a shortcut on its anti-correlated class (e.g.\ a Good actor in a negative review).

The dataset creation process involves identifying actor names in reviews -  through
a named entity recognition tagger - and using 
these to obtain a templated version of the review where actor names can be systematically replaced (see Appendix \ref{sec:actor_corr_appendix}). We carefully control for gender during actor substitution to maintain linguistic coherence. 
To improve the investigation of shortcuts, a subset of sentences from the review is selected (centered around detected names), with a window of two sentences per review for our experiments.
Not all reviews contain actor names, which is no problem for the training set which only injects shortcuts into a small selection of the reviews.

The dataset is divided into three splits: training, validation and test. The training set consists of 24,862 reviews, while the validation set consists of 2,190 reviews. 
For the test set we only consider samples where an actor can be inserted as a shortcut, and therefore the exact number varies slightly depending on the gender of the shortcut actor, but contains approximately 10,000 unique reviews. 
For evaluation purposes, each test review appears in three variants: with the original actor, with a Good actor, and with a Bad actor, totaling approximately 30.000 test instances.
Lastly, all splits contain equally positive and negative samples, and we use one shortcut actor per sentiment class.

\subsection{Experimental Setup}

We use the GPT2 model \cite{radford2019language} converting it to a classifier using the prompt template below. We make two modifications to the way we use the model output. Firstly, we only consider the output embedding of the last token stream. Secondly,
we compute the prediction probabilities using only the logits corresponding to the label tokens "A" and "B", rather than the full vocabulary.

\begin{figure}[h!]
\centering
\begin{BVerbatim}[fontsize=\small]
"Classify the sentiment of the movie review:
Review: """{review}"""

LABEL OPTIONS: A: negative  B: positive
LABEL:"
\end{BVerbatim}
\end{figure}

To inspect the effect of the shortcut we introduce the Anti-Correlated Accuracy Change (ACAC) which calculates the model's average drop in accuracy when anti-correlated shortcuts are inserted, compared to the original actor.
The ACAC is computed using the accuracy per subset as:
\begin{equation}
    \text{ACAC} = \frac{1}{2}\sum_{c \in {\text{Pos},\text{Neg}}} [\text{Acc}(X^c_{og}) - \text{Acc}(X^c_{ac})]
    \label{eq:acac_equation}
\end{equation}
Where $X^c_{j}$ is the subset of the test data which has class $c$ and actor name type $j \in \{og, ac\}$, which can be the original name ($og$), or the anti-correlated shortcut name ($ac$). And \textit{Acc($X_{j}^c$)} is the accuracy of this subset data.

\begin{figure*}[t] 
    \centering
    \subfloat[Test accuracy per category]{
        \adjustbox{valign=c}{
        \resizebox{0.28\textwidth}{!}{
         \begin{tabular}{@{}llll@{}}
             & \multicolumn{3}{c}{Actor class}\\
            \toprule
            Sentiment& Good& Original& Bad\\ \midrule 
            Positive& 96.78& 84.09& 54.30\\
            Negative& 33.43& 69.91& 87.41\end{tabular}    
        }
        }
        \label{tab:sentiment_class_test}
        }
    \subfloat[Shortcut Frequency]{
         \includegraphics[width=\imgwidthVar\textwidth]{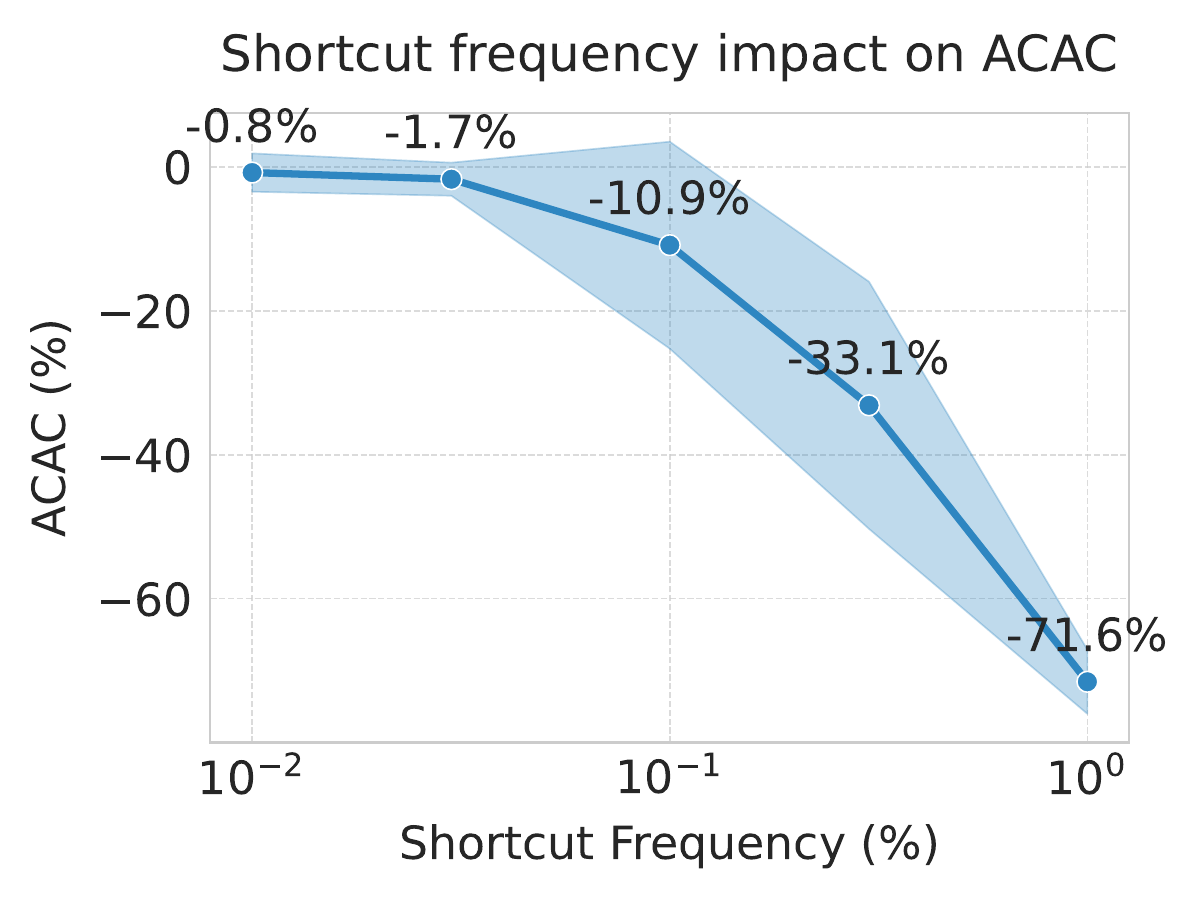}
        \label{fig:shortcut_freq}          
        }
        \subfloat[Shortcut Purity]{
         \includegraphics[width=\imgwidthVar\textwidth]{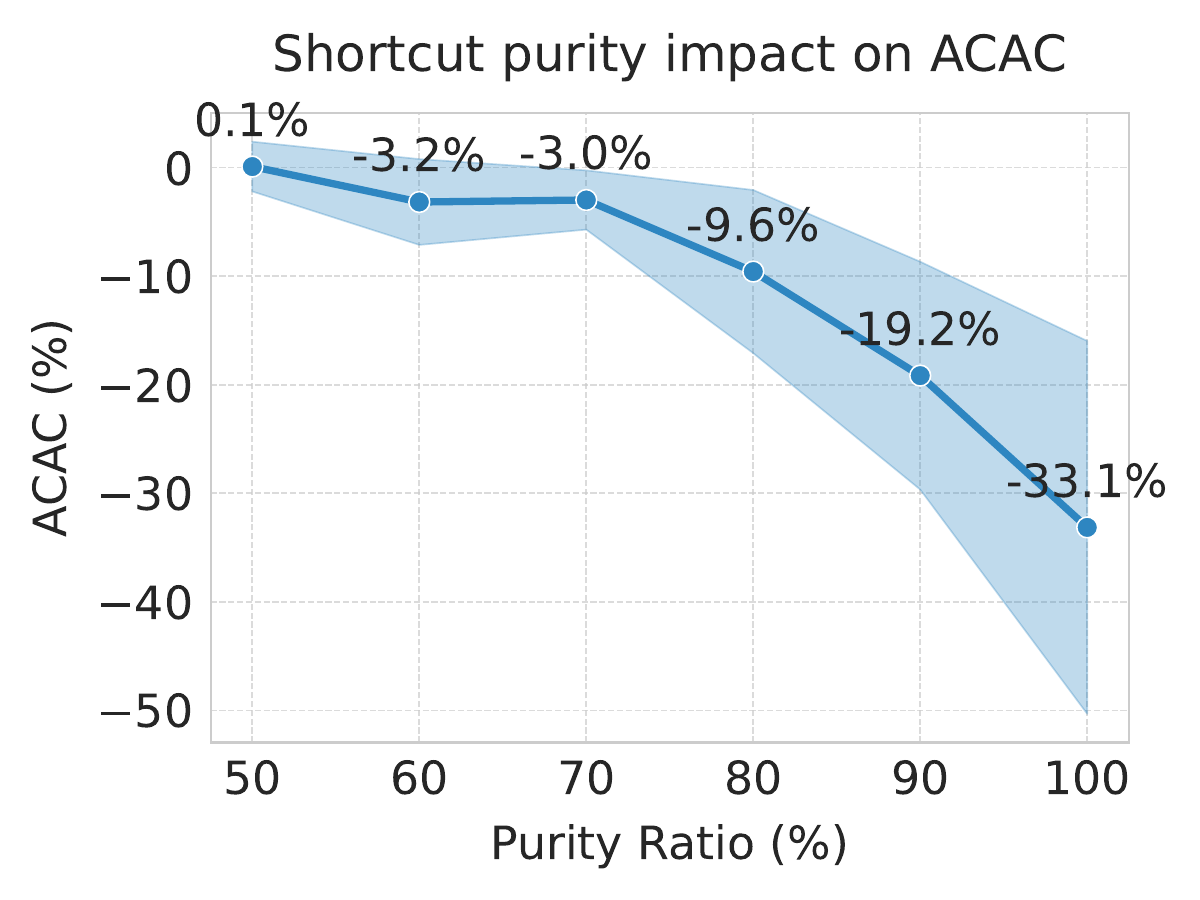}
        \label{fig:shortcut_purity}          
        }
    \caption{
    Effect of shortcuts on correlated and anti-correlated classes. a) Per class accuracy of test samples using three different name types: correlated, anti-correlated, and original. 
    b\&c) Effect of anti-correlated shortcuts (quantified by the ACAC metric of Equation \ref{eq:acac_equation}) when changing shortcut frequency (b) and purity ratio (c).}
    \label{fig:shortcut_performance_figures}
\end{figure*}

\subsection{Results}
We present the results in Figure \ref{fig:shortcut_performance_figures} as the mean over four different training instances (two times with male actors, and two times with female actors).

Table \ref{tab:sentiment_class_test} shows the accuracy per sentiment class using the three variants for each review, when trained using shortcuts in 0.3\% of the training set. 
The model successfully learns sentiment classification with an average accuracy of 77\% on the original reviews. The shortcuts significantly reduce this, causing an ACAC of 33\%.\footnote{The ACAC of the table in Figure \ref{tab:sentiment_class_test} is computed as $\frac{1}{2} [ (84.09 - 54.30) + (69.91 - 33.43)]  =   33.14$\%.}

In Figure \ref{fig:shortcut_freq}, we vary the shortcut percentage in the training data. 
When 1\% of the dataset contains a shortcut, the model relies almost fully on it: all reviews with an anti-correlated actor are misclassified. Moreover, a shortcut frequency of 0.1\% already has a significant impact.

Shortcuts will not always be absolute. We thus evaluate the impact of the purity of the shortcut. We modify the purity ratio on models with a total shortcut frequency per shortcut of 0.1\%. A purity ratio of 0.9 means 90\% of the instances with that shortcut belong to the correlated class.
Figure \ref{fig:shortcut_purity} shows that impure shortcut signals — that is, when the actor occasionally appears in both classes - also impact model behavior. A purity ratio of 80\% still leads to a substantial accuracy drop of 4\% on anti-correlated samples.

Unless stated otherwise, we use a shortcut frequency of 0.03\% (i.e.\ 72 reviews), with a purity ratio of 1.0 in the remainder of this paper.

\begin{figure*}[t]
     \centering
     \begin{subfigure}[b]{0.7\textwidth}
         \centering
         \includegraphics[width=\textwidth]{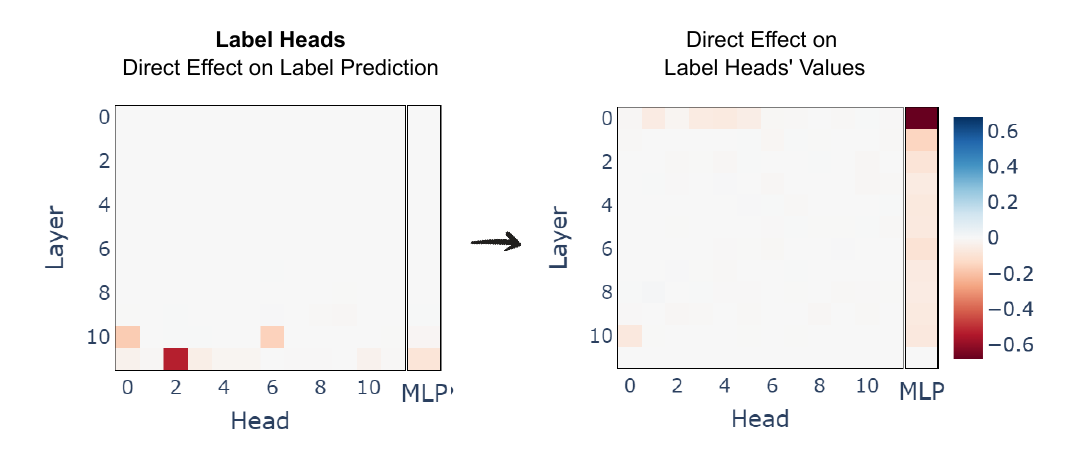}
         \caption{}
         \label{fig:patching_two_heatmaps}
     \end{subfigure}
     \begin{subfigure}[b]{0.29\textwidth}
         \centering
         \includegraphics[width=\textwidth]{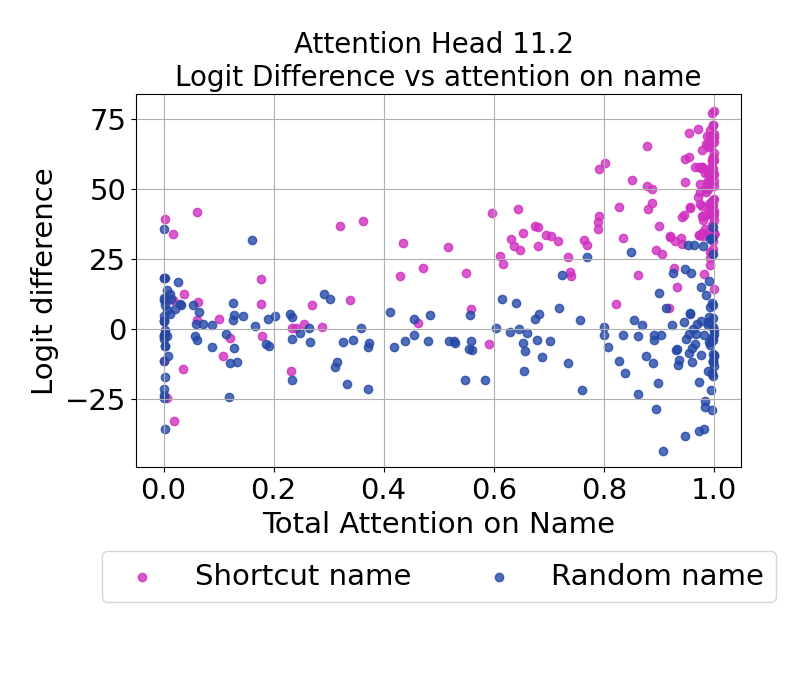}
         \caption{}         
          \label{fig:patching_scatter}
     \end{subfigure}
        \caption{Path Patching results on ActorCorr trained model for Bad actor in positive reviews. (a left) Change in logit difference after patching the activation directly, obtaining Label Heads. (a right) Change in logit difference after patching via Label Heads.
         (b) Cumulative attention on name tokens against the logit difference for Label Head 11.2
        }
        \label{fig:actorcorr_patching}
\end{figure*}

\section{How shortcuts are processed}\label{sec:shortcut_mechanism}

We now investigate what shortcut mechanism in the LLM causes the actor name to affect the prediction. 

\subsection{Experimental Setup}\label{sec:patching_experimental_setup}

Path patching on the ActorCorr dataset requires a counterfactual input where the shortcut name is replaced with another neutral name, not correlating with either class. The reference sentence $X$ and counterfactual sentence $\tilde{X}$ should contain the same number of tokens for efficient patching, therefore, we cannot simply use the original name for our counterfactual. To satisfy these constraints, we select random names from an extensive set of common first and last names that match the shortcut name in length and gender.

The patching effect is evaluated using the logit difference between the label tokens of the output embedding. Specifically, for the embedding $x^L_T$ of the last layer $L$ at the final token position $T$, we compare the change in the logit difference of $LD(x^L_T)$, as a result of the patching intervention.

We evaluate the effect of the Bad actor shortcut on the positive sentiment reviews and run path patching using 200 samples showing the mean results for one model. Appendix~\ref{sec:appendix_patching_imb_freq} provides the results for multiple runs showing the same general observations.

\subsection{Patching Results}\label{sec:patching_results}
Figure \ref{fig:patching_two_heatmaps} demonstrates the results of our shortcut circuit experiments, when patching the activations of the individual components (i.e. attention heads and MLPs).
The heatmap illustrates how specific attention heads are the most important contributors to the logits, mainly head 11.2 (i.e.\ layer 11, head 2), and to a lesser degree 10.10 and 10.6. 
Since the activation of these components directly affects the predicted class label, we refer to them as \textit{Label Heads}.
Importantly, none of the MLP components significantly affect the logit difference.

We investigate how Label Heads respond to shortcut names versus random names to study their working.
Figure \ref{fig:patching_scatter} shows that Label Head 11.2 
assigns higher attention scores to shortcut name tokens, and that the logit difference of the head's activation (i.e. $LD(a^{11,2}_T)$) is also greater for shortcuts compared to random names.

Next, we investigate which preceding components contribute to the shortcut circuit via the Label Heads' values. Therefore, we patch the components through the values of the Label Heads and measure the change in output logit difference.\footnote{Since the keys and values of the Label Heads both appeared relevant, we could patch via either. Appendix \ref{sec:appendix_keys_patching} shows that patching via the keys obtains similar components.}
Figure \ref{fig:patching_two_heatmaps} (right) reveals that mainly MLP layers are responsible. The first layer especially seems important, but many of the later MLP layers are doing something similar.

\paragraph{The Shortcut Mechanism}
Our patching experiments revealed that the shortcut circuit consisted of the first MLP layer and the Label Heads.
This connects to previous work, which demonstrated how attention heads are mainly responsible for moving information between token streams \cite{elhage2021mathematical}, while MLP layers function as dictionaries for knowledge retrieval \cite{geva2021transformer, meng2022locating}. Recent work has also found that early-layer MLPs can enrich entity, e.g. by finding related semantic attributes \cite{yu2024mechanistic, yu2023characterizing}.
Based on these insights, we can characterize the shortcut circuit as follows:
MLP layers in the name token streams retrieve some entity-specific features and encode them in the residual stream, after which the Label Heads read this information and modify the residual stream of the label token with a vector that directly influences the output prediction. 

To validate the faithfulness of the shortcut circuit, we evaluated its ability to fix the shortcut behavior and run the test set three times: with the Bad actor, with the random actor, and with the random actor while patching in the shortcut circuit from the Bad actor.
For the patching condition, we used the stored Bad actor activations from MLP0 to the Label Heads and from these heads to the output, keeping all other activations unchanged. Table \ref{tab:patching_faithfulness} demonstrates the circuit successfully reconstructed 57\% of the ACAC (11 / 19.5) for the anti-correlated class and 69\% (11.4 / 16.6) for the correlated class. It thus captures a significant portion of the model's shortcut behavior for both classification scenarios.

\begin{table}[h!]
\centering
\resizebox{\columnwidth}{!}{%
\begin{tabular}{l l l l} \hline 
 & Random& Bad& $\text{Random}_{\text{patch}}$\\ \hline 
Positive & 83.1& 63.5 (\textcolor{darkred}{-19.5})& 72.1 (\textcolor{darkred}{-11.0})\\ 
Negative & 72.2& 88.8 (\textcolor{lightblue}{+16.6})& 83.6 (\textcolor{lightblue}{+11.4})\\ 
\end{tabular}
}
\caption{Patching faithfulness result for the Bad actor on the two sentiment classes. Within brackets, accuracy changes with respect to random.}
\label{tab:patching_faithfulness}
\end{table}

\section{Classification via Feature Attribution}\label{sec:shortcut_classification}

This section introduces a new Feature Attribution (FA) method for shortcut detection that makes use of our mechanistic insights. We use existing FA methods as shortcut classifiers that generate per-word scores through sub-token aggregation as baselines. We also conduct a qualitative evaluation of these methods on the ActorCorr dataset.

\subsection{Feature Attribution Methods}

\paragraph{Head-based Token Attribution} Section~\ref{sec:shortcut_mechanism} revealed that shortcuts can change the attention pattern and the logit difference of the output activation of attention heads. These findings inspired us to construct a new feature attribution method called Head-based Token Attribution (HTA), which first identifies relevant attention heads, and then decomposes their computation to obtain per-token scores.

For the label token stream (indexed $T$), for each layer $l$ and head $h$, we compute the logit difference produced by that head's output activation $a^{l,h}_{T}$, which we denote as $LD(a^{l,h}_{T})$ (see Section~\ref{MI}). Heads exceeding an absolute logit difference with a threshold $\tau$ are selected for the final computation, where $\mathcal{H}$ contains these head indices (l,h).\footnote{Parameter $\tau$ reduces the search space with limited performance impact, as ignored heads have low logit differences and minimally contribute to the final score.}
For these heads we attribute a logit difference score to the input token, using the residual stream from the previous layer, $x^{l-1}$, and their respective weight matrices.
From these values we compute $A^{l,h}_{T, i}$ which represents the attention pattern over the input tokens for destination token $T$, while the VO matrix ($W^{l,h}_{VO} $) tells us how the embeddings would be modified by this head during attention.
HTA thus decomposes the head's computation. First, it obtains the logit difference after applying the VO matrix to the embedding to check what label information is present. Then it multiplies it by the attention score, to gather how much of it would be moved by the attention head. The final HTA score per input token is the result of summing the results for the earlier found top heads $\mathcal{H}$.

\begin{equation}
    \text{HTA}(x^0_i) = \sum_{(l,h) \in \mathcal{H}}  A^{l,h}_{T, i} \cdot LD(x^{l-1}_i W^{l,h}_{VO} ) 
\end{equation}

\paragraph{Baseline Methods}
We compare HTA against two established feature attribution methods: Integrated Gradients (IG)~\cite{sundararajan2017axiomatic}, a gradient-based approach that integrates attribution along a linear path from a baseline to the input, and LIME~\cite{ribeiro2016should}, a model-agnostic method that fits an interpretable local model via input permutations. See Appendix~\ref{sec:appendix_fa_details} for details.

\begin{figure*}[t]
    \centering
    \subfloat[AUROC Metric]{
         \includegraphics[height=4cm]{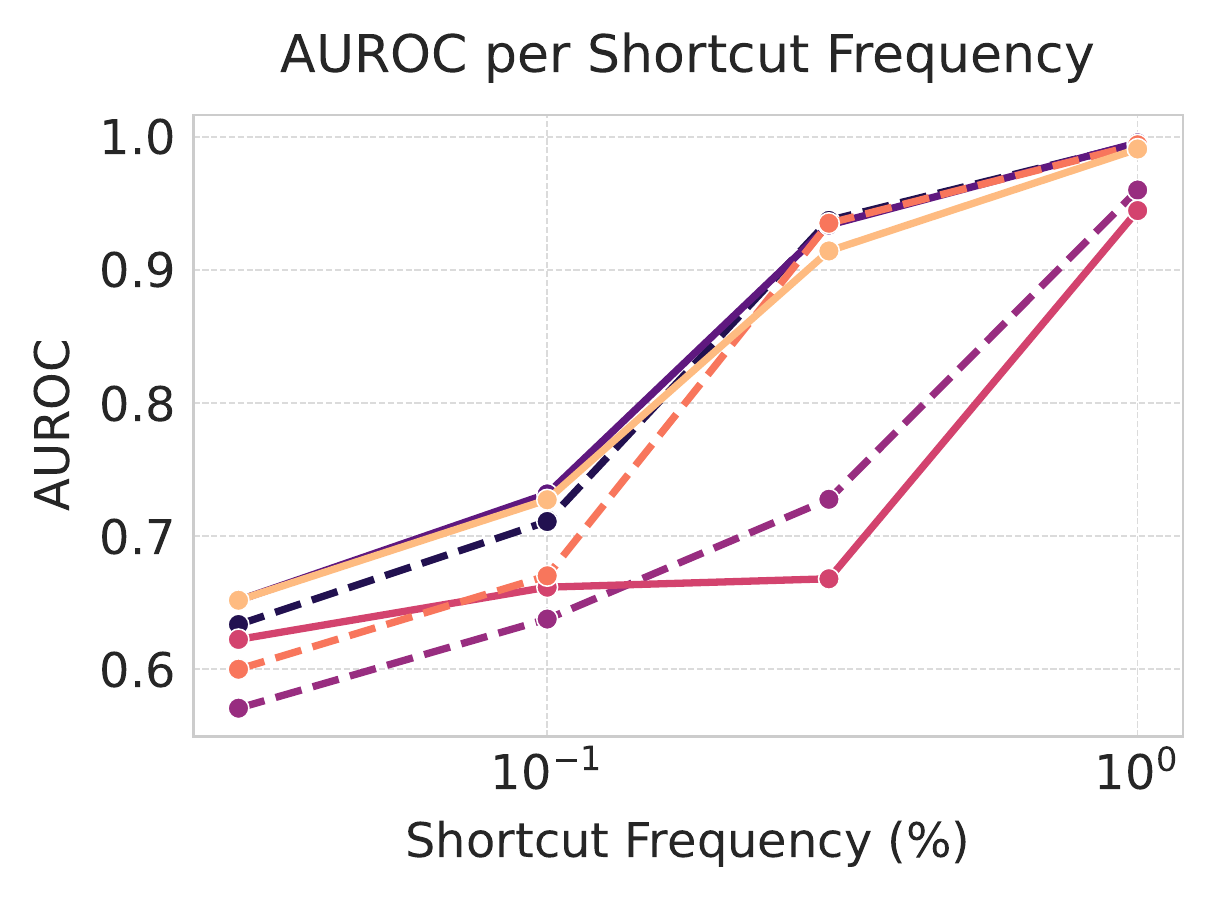}
        \label{fig:roc_auc_results}          
        }
        \subfloat[Cohen's d Metric]{
         \includegraphics[height=4cm]{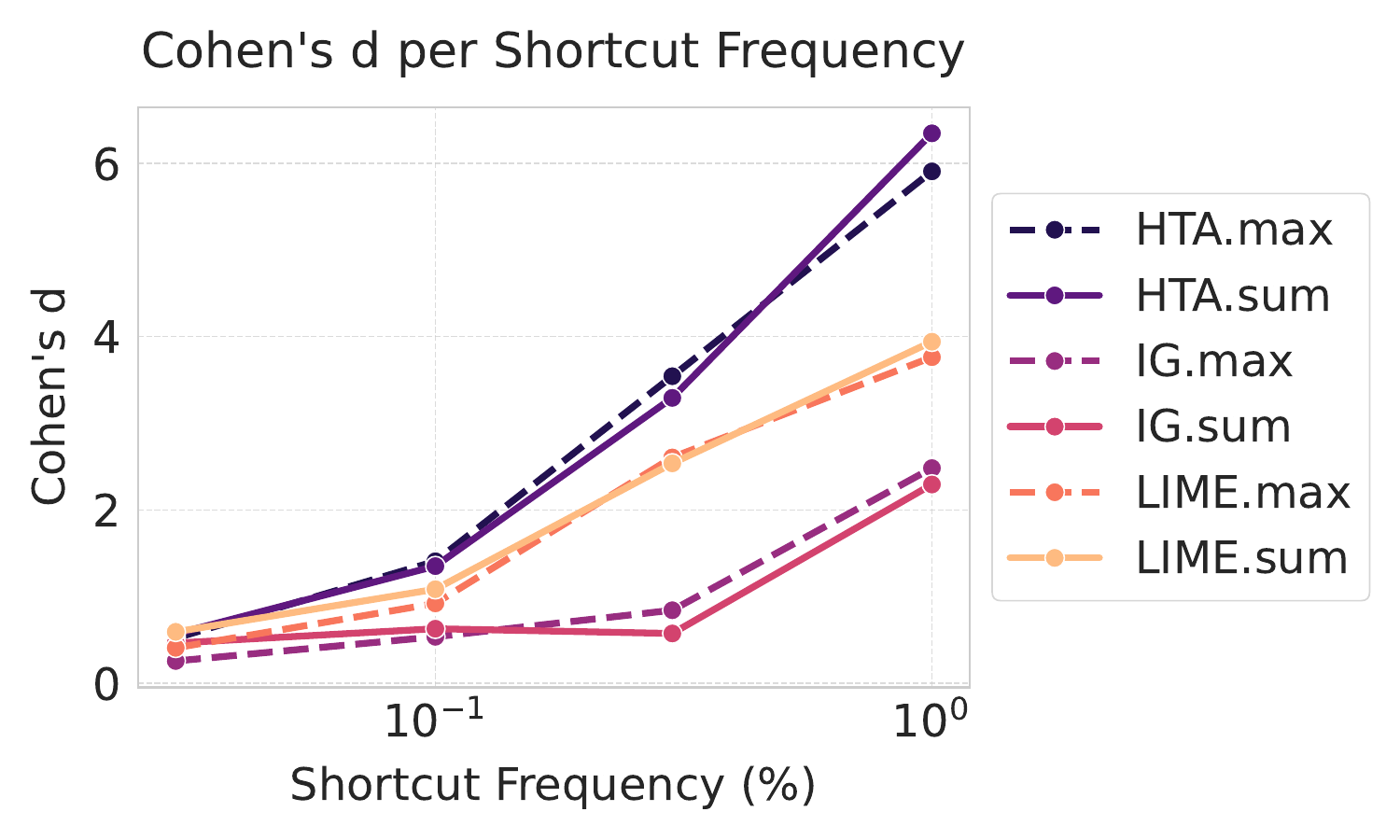}
        \label{fig:cohensd_results}          
        }
        \subfloat[Score Distributions]{
         \includegraphics[height=4cm]{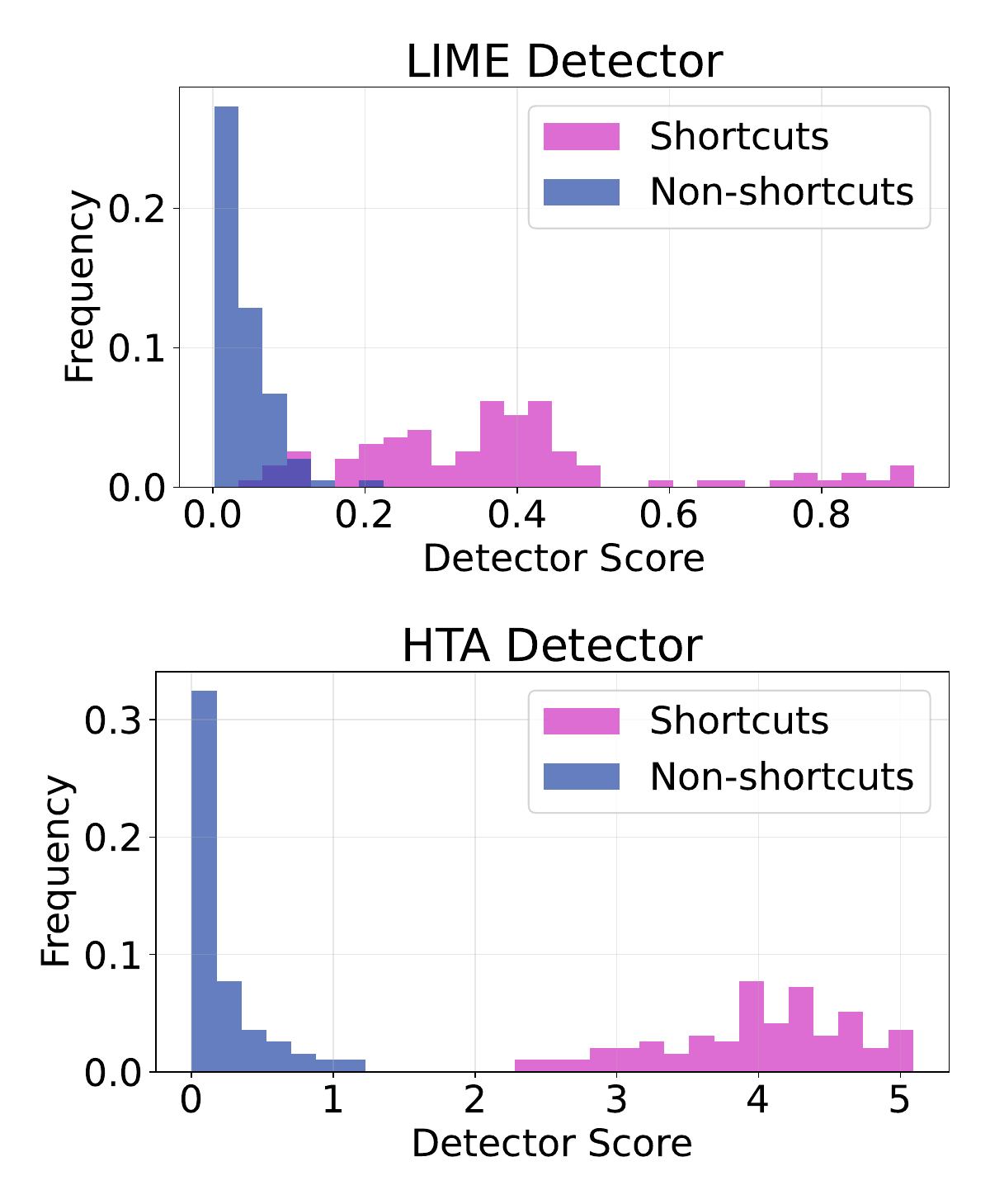}
        \label{fig:score_distributions_example}          
        }
    \caption{a,b) Shortcut classification evaluated via distribution separation metrics for the three feature attribution methods HTA, LIME and IG, using the two aggregation functions (max, sum). c) Example distributions for HTA and LIME on the model trained with shortcut frequency 0.003. }
    \label{fig:separation_metric_results}
\end{figure*}

\subsection{Experimental Setup}\label{sec:shortcut_detect_via_attrib}
We implement the feature attribution methods as shortcut classifiers using their importance scores per token.
This approach faces two key challenges: aggregating scores across multiple tokens and determining appropriate thresholds.
Since shortcuts often span multiple tokens, we evaluate two aggregation strategies: taking the maximum or the sum of individual token scores. Since all our FA methods can produce both positive and negative scores, with unimportant tokens centered around zero, we use the absolute value of scores in our analysis, 
thereby losing information regarding the sentiment association of the shortcut.

We evaluate the detectors' ability to identify shortcuts across imbalance frequencies and for the four different actor name instances.
We (again) focus on the effect of the Bad actor on the positive reviews.
We randomly select 1000 unique positive reviews for each test set, where each review undergoes two evaluations: one with the Bad actor and one with the random actor (same as Section \ref{sec:patching_experimental_setup}). %
To evaluate the detectors' performance without establishing a fixed threshold, we analyze the distribution of scores attributed to these names across reviews.

\paragraph{Classification Evaluation Metrics}
To measure the separability in score distributions between shortcut and non-shortcut names, we use two metrics provide complementary insights in separability.
The Area Under the ROC curve (AUROC) \cite{bradley1997use} provides a measure of overlap between the two distributions, with 1.0 indicating perfect separability. 
Since practical applications may require threshold estimation from limited samples, we also compute Cohen's d~\cite{cohen1988statistical}:
\begin{equation}
\text{Cohen's d} = \frac{\mu_{1}-\mu_{2} }{\sigma_{\text{pool}}}
\end{equation}
Here $\sigma_{pool}$ is the pooled standard deviation between the two distributions, and is formally defined as $\sigma_{\text{pool}} = \sqrt{(\sigma_1^2 + \sigma_2^2)/2}$. Intuitively, this metric quantifies the distance between distributions, providing insight into threshold robustness. Figure~\ref{fig:distribution_metrics} illustrates how these metrics capture different aspects of distribution separation.

\subsection{Shortcut Classification results}
Figure \ref{fig:separation_metric_results} demonstrates the various performance characteristics in shortcut detection capabilities.
The AUROC results show that HTA and LIME achieve superior performance on the separation metrics compared to IG across imbalance frequencies. 
Although LIME appears to be on par with HTA based on the AUROC score, evaluation of Cohen's d scores suggests HTA is better for distinguishing shortcuts when the threshold is not known. 
To illustrate these differences better, we evaluate the score distributions using max-aggregation for the model used in our patching evaluation, with
shortcut frequency 0.3\%.
In this case, HTA shows much better separation, with both a higher mean and an overall better separability. 
The choice of aggregation method seems to have a varying but minor effect, where \textit{sum} works well for most HTA cases, but for LIME and IG \textit{max} might be better depending on the shortcut frequency.

Computationally, HTA is much more efficient than the other two methods, requiring only one forward pass and no gradients, compared to 3000 perturbed forward passes of LIME and the compute-intensive path-integrated gradient technique of IG.

\section{Shortcut Mitigation}\label{sec:shortcut_mitigation}

HTA can thus identify shortcuts and find how they are processed. This offers a potential mitigation strategy: Since attention heads $\mathcal{H}$ producing high logit-differences focus mostly on name tokens, selective head ablation may be an effective remedy.

 \begin{table}[h!]
    \centering
    \resizebox{\columnwidth}{!}{%
    \begin{tabular}{@{}llll@{}}
        & \multicolumn{3}{c}{Actor class}\\
        \toprule
        Class & Good & Original& Bad\\ \midrule
        Pos& 89.4 (\textcolor{darkred}{-8.3}) & 82.2 (\textcolor{darkred}{-0.3}) & 81.4 (\textcolor{lightblue}{+18.5})\\
        Neg& 61.8 (\textcolor{lightblue}{+30.2}) & 73.1 (\textcolor{lightblue}{+0.6}) & 74.8 (\textcolor{darkred}{-13.9})\\
    \end{tabular}
    }
    \caption{Test accuracy after Label Heads ablation. Brackets show difference from non-ablated model.}
    \label{tab:ablate_head_results}
\end{table}

Experimental results, presented in Table~\ref{tab:ablate_head_results}, demonstrate that ablating these heads significantly reduces the shortcut effects. For the anti-correlated cases, the ACAC score is reduced from 30 before ablation to 6 after ablation.
However, later layer heads can compensate for the behavior of ablated attention heads \cite{mcgrath2023hydra}. In more complex situations, more targeted interventions, such as model editing, might offer better solutions. %

\section{Qualitative Analysis}\label{sec:qualitative_analysis_main}
To understand HTA's broader applicability, we analyze its attribution scores on reviews without our inserted shortcuts and compare against LIME and IG, see Appendix \ref{sec:appendix_qualitative_analysis} for the full analysis and results. Our analysis reveals three key characteristics of HTA. Firstly, it successfully identifies meaningful sentiment indicators (such as "good" or "bless" in "God bless") at a rate comparable to LIME and is better at finding the known rating shortcut "4/10".
Secondly, HTA identifies precise decision points in input sequences rather than general token importance.
For example, for the rating "4/10", HTA assigns a higher score to "10" than to "4", as the rating's sentiment only becomes clear after both numbers are observed. 
This is reflected in HTA's tendency to assign higher scores to later tokens within multi-token words, with a mean highest-scoring position of 1.69 versus 1.60 and 1.51 for LIME and IG.
Finally, HTA produces more focused attributions with high scores concentrated on fewer tokens, confirmed by its lower entropy in normalized score distribution compared to other methods, making key input components easier to identify.

\section{Conclusion}
We investigated the mechanisms that process shortcuts in LLMs, specifically focusing on the spurious correlation of actor names in movie reviews. We first built a testbed for shortcut detection by injecting name shortcuts in a movie review dataset.
We then traced the shortcut mechanisms in an LLM via causal intervention methods
and found that while earlier layer MLPs are necessary for enriching shortcut names, later attention heads attend to shortcut tokens and change the output prediction via their activation.
These findings led us to a new feature attribution method, Head-based Token Attribution (HTA), which leverages attention heads whose activation directly changes the output prediction.
Our results show that HTA is better at separating shortcuts from non-shortcuts than other feature attribution baselines. Our findings using HTA confirm that the model begins generating predictions at intermediate input stages, effectively reaching conclusions before processing the full context.

\section*{Limitations}

Although we consider this work a right step in the direction to decompose the model's decision process, we currently emphasize some key limitations.

Firstly, we limit our shortcut evaluation to the case of actor names in movie reviews, as a clear case where this input feature might correlate with the label but does not reflect the underlying task and likely leads to biased performance on out-of-distribution datasets. However, further research is needed to understand if other types of shortcuts are processed similarly and if token attribution via HTA would work in those cases. 

Secondly, we limit our experiments to Transformer decoder models.
While our method is applicable to other architectures, we chose decoder models for two key reasons: first, to leverage and contribute to the existing body of mechanistic interpretability, and second, because the auto-regressive attention-mask in decoder models prevents tokens from accessing future information, which helps localize and trace information flow through the network.

While our causal intervention results in Section \ref{sec:shortcut_mechanism} find a clear causal relation in the case of name shortcut, further research is needed to determine if our Head-based Token Attribution offers reliable attribution of shortcuts in other situations. 
Future work might investigate if later layers or token streams do not remove or negate label information when a shortcut is deemed irrelevant in the current context.  

Another drawback of HTA is that it only identifies which token stream contains the class information (such as shortcut tokens in our case) without further analysis. If the model properly processes a sentence contextually rather than using shortcuts, the class information might be stored in the final token stream (e.g., a period "."). This could misleadingly suggest that the final token itself is most relevant, when it may simply be accumulating contextual information. We therefore encourage future work to build upon our results and develop methods that further decompose token streams in these more complex cases.

\section*{Ethics Statement}
Our work contributes to the existing body of literature that aims to decompose the computations in LLMs, which is crucial for safe deployment of these AI systems.
Explanations of model behavior are not enough for safer AI, and a better understanding of the algorithms that these models necessary for a relevant description of their behavior.

\section*{Acknowledgments}
This research was partially funded by the Hybrid Intelligence Center, a 10-year program funded by the Dutch Ministry of Education, Culture and Science through the Netherlands Organisation for Scientific Research, \url{https://hybrid-intelligence-centre.nl}. We would like to thank Martin Carrasco Castaneda, Jonathan Kamp, Urja Khurana, Pia Sommerauer, and Sergey Troshin for their valuable feedback.

\bibliography{custom}

\begin{thebibliography}{37}
\expandafter\ifx\csname natexlab\endcsname\relax\def\natexlab#1{#1}\fi

\bibitem[{Alzantot et~al.(2018)Alzantot, Sharma, Elgohary, Ho, Srivastava, and Chang}]{alzantot-etal-2018-generating}
Moustafa Alzantot, Yash Sharma, Ahmed Elgohary, Bo-Jhang Ho, Mani Srivastava, and Kai-Wei Chang. 2018.
\newblock \href {https://doi.org/10.18653/v1/D18-1316} {Generating natural language adversarial examples}.
\newblock In \emph{Proceedings of the 2018 Conference on Empirical Methods in Natural Language Processing}, pages 2890--2896, Brussels, Belgium. Association for Computational Linguistics.

\bibitem[{Bai et~al.(2021)Bai, Liang, Zhang, Li, Bai, and Wang}]{bai2021attentions}
Bing Bai, Jian Liang, Guanhua Zhang, Hao Li, Kun Bai, and Fei Wang. 2021.
\newblock Why attentions may not be interpretable?
\newblock In \emph{Proceedings of the 27th ACM SIGKDD conference on knowledge discovery \& data mining}, pages 25--34.

\bibitem[{Bastings et~al.(2022)Bastings, Ebert, Zablotskaia, Sandholm, and Filippova}]{bastings2022will}
Jasmijn Bastings, Sebastian Ebert, Polina Zablotskaia, Anders Sandholm, and Katja Filippova. 2022.
\newblock “will you find these shortcuts?” a protocol for evaluating the faithfulness of input salience methods for text classification.
\newblock In \emph{Proceedings of the 2022 Conference on Empirical Methods in Natural Language Processing}, pages 976--991.

\bibitem[{Bradley(1997)}]{bradley1997use}
Andrew~P Bradley. 1997.
\newblock The use of the area under the roc curve in the evaluation of machine learning algorithms.
\newblock \emph{Pattern recognition}, 30(7):1145--1159.

\bibitem[{Cohen(1988)}]{cohen1988statistical}
J~Cohen. 1988.
\newblock Statistical power analysis for the behavioral sciences . hillsdale, nj: Eribaum.

\bibitem[{Du et~al.(2023)Du, He, Zou, Tao, and Hu}]{du2023shortcut}
Mengnan Du, Fengxiang He, Na~Zou, Dacheng Tao, and Xia Hu. 2023.
\newblock Shortcut learning of large language models in natural language understanding.
\newblock \emph{Communications of the ACM}, 67(1):110--120.

\bibitem[{Du et~al.(2021)Du, Manjunatha, Jain, Deshpande, Dernoncourt, Gu, Sun, and Hu}]{du-etal-2021-towards}
Mengnan Du, Varun Manjunatha, Rajiv Jain, Ruchi Deshpande, Franck Dernoncourt, Jiuxiang Gu, Tong Sun, and Xia Hu. 2021.
\newblock \href {https://doi.org/10.18653/v1/2021.naacl-main.71} {Towards interpreting and mitigating shortcut learning behavior of {NLU} models}.
\newblock In \emph{Proceedings of the 2021 Conference of the North American Chapter of the Association for Computational Linguistics: Human Language Technologies}, pages 915--929, Online. Association for Computational Linguistics.

\bibitem[{Elhage et~al.(2021)Elhage, Nanda, Olsson, Henighan, Joseph, Mann, Askell, Bai, Chen, Conerly et~al.}]{elhage2021mathematical}
Nelson Elhage, Neel Nanda, Catherine Olsson, Tom Henighan, Nicholas Joseph, Ben Mann, Amanda Askell, Yuntao Bai, Anna Chen, Tom Conerly, et~al. 2021.
\newblock A mathematical framework for transformer circuits.
\newblock \emph{Transformer Circuits Thread}, 1(1):12.

\bibitem[{Friedman et~al.(2022)Friedman, Wettig, and Chen}]{friedman2022finding}
Dan Friedman, Alexander Wettig, and Danqi Chen. 2022.
\newblock Finding dataset shortcuts with grammar induction.
\newblock In \emph{Proceedings of the 2022 Conference on Empirical Methods in Natural Language Processing}, pages 4345--4363.

\bibitem[{Geiger et~al.(2021)Geiger, Lu, Icard, and Potts}]{geiger2021causal}
Atticus Geiger, Hanson Lu, Thomas Icard, and Christopher Potts. 2021.
\newblock Causal abstractions of neural networks.
\newblock \emph{Advances in Neural Information Processing Systems}, 34:9574--9586.

\bibitem[{Geva et~al.(2023)Geva, Bastings, Filippova, and Globerson}]{geva2023dissecting}
Mor Geva, Jasmijn Bastings, Katja Filippova, and Amir Globerson. 2023.
\newblock Dissecting recall of factual associations in auto-regressive language models.
\newblock In \emph{Proceedings of the 2023 Conference on Empirical Methods in Natural Language Processing}, pages 12216--12235.

\bibitem[{Geva et~al.(2021)Geva, Schuster, Berant, and Levy}]{geva2021transformer}
Mor Geva, Roei Schuster, Jonathan Berant, and Omer Levy. 2021.
\newblock Transformer feed-forward layers are key-value memories.
\newblock In \emph{Proceedings of the 2021 Conference on Empirical Methods in Natural Language Processing}, pages 5484--5495.

\bibitem[{Hanna et~al.(2024)Hanna, Liu, and Variengien}]{hanna2024does}
Michael Hanna, Ollie Liu, and Alexandre Variengien. 2024.
\newblock How does gpt-2 compute greater-than?: Interpreting mathematical abilities in a pre-trained language model.
\newblock \emph{Advances in Neural Information Processing Systems}, 36.

\bibitem[{Kamp et~al.(2024)Kamp, Beinborn, and Fokkens}]{kamp2024role}
Jonathan Kamp, Lisa Beinborn, and Antske Fokkens. 2024.
\newblock The role of syntactic span preferences in post-hoc explanation disagreement.
\newblock In \emph{Proceedings of the 2024 Joint International Conference on Computational Linguistics, Language Resources and Evaluation (LREC-COLING 2024)}, pages 16066--16078.

\bibitem[{Maas et~al.(2011)Maas, Daly, Pham, Huang, Ng, and Potts}]{imdb_dataset}
Andrew~L. Maas, Raymond~E. Daly, Peter~T. Pham, Dan Huang, Andrew~Y. Ng, and Christopher Potts. 2011.
\newblock \href {http://www.aclweb.org/anthology/P11-1015} {Learning word vectors for sentiment analysis}.
\newblock In \emph{Proceedings of the 49th Annual Meeting of the Association for Computational Linguistics: Human Language Technologies}, pages 142--150, Portland, Oregon, USA. Association for Computational Linguistics.

\bibitem[{Madsen et~al.(2022)Madsen, Reddy, and Chandar}]{madsen2022post}
Andreas Madsen, Siva Reddy, and Sarath Chandar. 2022.
\newblock Post-hoc interpretability for neural nlp: A survey.
\newblock \emph{ACM Computing Surveys}, 55(8):1--42.

\bibitem[{McGrath et~al.(2023)McGrath, Rahtz, Kramar, Mikulik, and Legg}]{mcgrath2023hydra}
Thomas McGrath, Matthew Rahtz, Janos Kramar, Vladimir Mikulik, and Shane Legg. 2023.
\newblock The hydra effect: Emergent self-repair in language model computations.
\newblock \emph{arXiv preprint arXiv:2307.15771}.

\bibitem[{Meng et~al.(2022)Meng, Bau, Andonian, and Belinkov}]{meng2022locating}
Kevin Meng, David Bau, Alex Andonian, and Yonatan Belinkov. 2022.
\newblock Locating and editing factual associations in gpt.
\newblock \emph{Advances in Neural Information Processing Systems}, 35:17359--17372.

\bibitem[{Naik et~al.(2018)Naik, Ravichander, Sadeh, Rose, and Neubig}]{naik-etal-2018-stress}
Aakanksha Naik, Abhilasha Ravichander, Norman Sadeh, Carolyn Rose, and Graham Neubig. 2018.
\newblock \href {https://aclanthology.org/C18-1198} {Stress test evaluation for natural language inference}.
\newblock In \emph{Proceedings of the 27th International Conference on Computational Linguistics}, pages 2340--2353, Santa Fe, New Mexico, USA. Association for Computational Linguistics.

\bibitem[{Nostalgebraist(2020)}]{nostalgebraist2020logitlens}
Nostalgebraist. 2020.
\newblock \href {https://www.lesswrong.com/posts/AcKRB8wDpdaN6v6ru/interpreting-gpt-the-logit-lens} {interpreting gpt: the logit lens}.
\newblock \emph{LessWrong}.

\bibitem[{Olah et~al.(2020)Olah, Cammarata, Schubert, Goh, Petrov, and Carter}]{olah2020zoom}
Chris Olah, Nick Cammarata, Ludwig Schubert, Gabriel Goh, Michael Petrov, and Shan Carter. 2020.
\newblock Zoom in: An introduction to circuits.
\newblock \emph{Distill}, 5(3):e00024--001.

\bibitem[{Pearl(2009)}]{pearl2009causality}
Judea Pearl. 2009.
\newblock \emph{Causality}.
\newblock Cambridge university press.

\bibitem[{Pezeshkpour et~al.(2022)Pezeshkpour, Jain, Singh, and Wallace}]{pezeshkpour-etal-2022-combining}
Pouya Pezeshkpour, Sarthak Jain, Sameer Singh, and Byron Wallace. 2022.
\newblock \href {https://doi.org/10.18653/v1/2022.findings-acl.153} {Combining feature and instance attribution to detect artifacts}.
\newblock In \emph{Findings of the Association for Computational Linguistics: ACL 2022}, pages 1934--1946, Dublin, Ireland. Association for Computational Linguistics.

\bibitem[{Pezeshkpour et~al.(2021)Pezeshkpour, Jain, Wallace, and Singh}]{pezeshkpour-etal-2021-empirical}
Pouya Pezeshkpour, Sarthak Jain, Byron Wallace, and Sameer Singh. 2021.
\newblock \href {https://doi.org/10.18653/v1/2021.naacl-main.75} {An empirical comparison of instance attribution methods for {NLP}}.
\newblock In \emph{Proceedings of the 2021 Conference of the North American Chapter of the Association for Computational Linguistics: Human Language Technologies}, pages 967--975, Online. Association for Computational Linguistics.

\bibitem[{Radford et~al.(2019)Radford, Wu, Child, Luan, Amodei, Sutskever et~al.}]{radford2019language}
Alec Radford, Jeffrey Wu, Rewon Child, David Luan, Dario Amodei, Ilya Sutskever, et~al. 2019.
\newblock Language models are unsupervised multitask learners.
\newblock \emph{OpenAI blog}, 1(8):9.

\bibitem[{R{\"a}uker et~al.(2023)R{\"a}uker, Ho, Casper, and Hadfield-Menell}]{rauker2023toward}
Tilman R{\"a}uker, Anson Ho, Stephen Casper, and Dylan Hadfield-Menell. 2023.
\newblock Toward transparent ai: A survey on interpreting the inner structures of deep neural networks.
\newblock In \emph{2023 ieee conference on secure and trustworthy machine learning (satml)}, pages 464--483. IEEE.

\bibitem[{Ribeiro et~al.(2016)Ribeiro, Singh, and Guestrin}]{ribeiro2016should}
Marco~Tulio Ribeiro, Sameer Singh, and Carlos Guestrin. 2016.
\newblock " why should i trust you?" explaining the predictions of any classifier.
\newblock In \emph{Proceedings of the 22nd ACM SIGKDD international conference on knowledge discovery and data mining}, pages 1135--1144.

\bibitem[{Ribeiro et~al.(2020)Ribeiro, Wu, Guestrin, and Singh}]{ribeiro2020beyond}
Marco~Tulio Ribeiro, Tongshuang Wu, Carlos Guestrin, and Sameer Singh. 2020.
\newblock Beyond accuracy: Behavioral testing of nlp models with checklist.
\newblock In \emph{Proceedings of the 58th Annual Meeting of the Association for Computational Linguistics}, pages 4902--4912.

\bibitem[{Ross et~al.(2021)Ross, Marasovi{\'c}, and Peters}]{ross-etal-2021-explaining}
Alexis Ross, Ana Marasovi{\'c}, and Matthew Peters. 2021.
\newblock \href {https://doi.org/10.18653/v1/2021.findings-acl.336} {Explaining {NLP} models via minimal contrastive editing ({M}i{CE})}.
\newblock In \emph{Findings of the Association for Computational Linguistics: ACL-IJCNLP 2021}, pages 3840--3852, Online. Association for Computational Linguistics.

\bibitem[{Sundararajan et~al.(2017)Sundararajan, Taly, and Yan}]{sundararajan2017axiomatic}
Mukund Sundararajan, Ankur Taly, and Qiqi Yan. 2017.
\newblock Axiomatic attribution for deep networks.
\newblock In \emph{International conference on machine learning}, pages 3319--3328. PMLR.

\bibitem[{Vaswani et~al.(2017)Vaswani, Shazeer, Parmar, Uszkoreit, Jones, Gomez, Kaiser, and Polosukhin}]{vaswani2017attention}
Ashish Vaswani, Noam Shazeer, Niki Parmar, Jakob Uszkoreit, Llion Jones, Aidan~N Gomez, {\L}ukasz Kaiser, and Illia Polosukhin. 2017.
\newblock Attention is all you need.
\newblock \emph{Advances in neural information processing systems}, 30.

\bibitem[{Vig et~al.(2020)Vig, Gehrmann, Belinkov, Qian, Nevo, Singer, and Shieber}]{vig2020investigating}
Jesse Vig, Sebastian Gehrmann, Yonatan Belinkov, Sharon Qian, Daniel Nevo, Yaron Singer, and Stuart Shieber. 2020.
\newblock Investigating gender bias in language models using causal mediation analysis.
\newblock \emph{Advances in neural information processing systems}, 33:12388--12401.

\bibitem[{Wang et~al.(2023)Wang, Variengien, Conmy, Shlegeris, and Steinhardt}]{wang2022interpretability}
Kevin~Ro Wang, Alexandre Variengien, Arthur Conmy, Buck Shlegeris, and Jacob Steinhardt. 2023.
\newblock Interpretability in the wild: a circuit for indirect object identification in gpt-2 small.
\newblock In \emph{The Eleventh International Conference on Learning Representations}.

\bibitem[{Wang et~al.(2022)Wang, Sridhar, Yang, and Wang}]{wang2022identifying}
Tianlu Wang, Rohit Sridhar, Diyi Yang, and Xuezhi Wang. 2022.
\newblock Identifying and mitigating spurious correlations for improving robustness in nlp models.
\newblock In \emph{Findings of the Association for Computational Linguistics: NAACL 2022}, pages 1719--1729.

\bibitem[{Yu et~al.(2024)Yu, Cao, Cheung, and Dong}]{yu2024mechanistic}
Lei Yu, Meng Cao, Jackie Chi~Kit Cheung, and Yue Dong. 2024.
\newblock Mechanistic understanding and mitigation of language model non-factual hallucinations.
\newblock In \emph{Findings of the Association for Computational Linguistics: EMNLP 2024}, pages 7943--7956.

\bibitem[{Yu et~al.(2023)Yu, Merullo, and Pavlick}]{yu2023characterizing}
Qinan Yu, Jack Merullo, and Ellie Pavlick. 2023.
\newblock Characterizing mechanisms for factual recall in language models.
\newblock In \emph{Proceedings of the 2023 Conference on Empirical Methods in Natural Language Processing}, pages 9924--9959.

\bibitem[{Zhang and Nanda(2023)}]{zhang2023towards}
Fred Zhang and Neel Nanda. 2023.
\newblock Towards best practices of activation patching in language models: Metrics and methods.
\newblock In \emph{The Twelfth International Conference on Learning Representations}.

\end{thebibliography}

\bibliographystyle{acl_natbib}

\appendix

\section{Appendix - Formalization}
\label{sec:appendix}

\subsection{Transformer Formalization}\label{sec:apx_transformer_notation_full}

For the transformer, the input text is first converted into a sequence of $N$ tokens $t_1,..., t_N$. Each token $t_i$ is then transformed into an
embedding $x_i$ of size $d_{resid}$ using the embedding matrix $W_e \in \mathbb{R}^{|V| \times d_{resid}}$, where $|V|$ is the size of the vocabulary.
Leading to the sequence of embeddings 
, $X^0  \in \mathbb{R}^{N \times d}$, where $0$ refers to the 0th layer or input layer.

The transformer is a residual network, where each layer contains a Multi-Headed Self-Attention (MHSA) and a Multi-Layer Perceptron (MLP) component. The connection from the input embedding to the output embedding to which these components add their embedding, or activation, is called the \textit{residual stream}. 
Formally, the attention activation is firstly computed as $a^l = MHSA(X^l)$, after which the MLP activation is computed as $m^l = MLP(X^l + a^l)$, resulting in the new residual embeddings:
\begin{equation}
X^{l+1}  = X^l + m^l + a^l
\end{equation}
After the last layer the final embeddings are projected to a vector of size $|V|$, using the unembed matrix $W_{u} \in \mathbb{R}^{d_{resid} \times |V|}$ to obtain the logits for each embedding. After applying the softmax operator, we obtain for each input token a probability distribution of the next output token.
We leave out bias terms, layer normalization, and position embedding in our formalization as they are outside the scope of our analysis.

\paragraph{Attention Heads}
Following \newcite{elhage2021mathematical}, the activation of the MHSA $a^l$ can be further decomposed as the sum of each attention head's contribution.
Each attention head contains the weight matrices $W_K, W_Q, W_V \in \mathbb{R}^{d_{resid} \times d_k}$, to compute the key, query, and value vectors.  
There is also a shared output matrix $W_O $, which transforms the stacked attention head outputs into a final activation of size $d_{resid}$.
Following \newcite{elhage2021mathematical}, the output matrix can be decomposed by selecting the columns that would match the specific attention head, resulting in $W^{l,h}_O \in \mathbb{R}^{d_k \times d_{resid} }$. Additionally, the output and value matrices can be reduced to a single matrix $W^{l,h}_{VO} = W^{l,h}_V W^{l,h}_O$, so that $W^{l,h}_{VO} \in \mathbb{R}^{d_{resid} \times d_{resid}}$.

The keys and queries are used to compute the attention score from the source token to each destination token, $A^{l,h}_{s,d}$, so that $A^{l,h} \in \mathbb{R}^{N \times N}$, but for the decoder a lower triangle mask is applied so that each token cannot attend to tokens after it.

\begin{equation}
    a^{l, h} =(A^{l, h} \cdot X^{l}  W_v^{l,h} )  W_o^{l,h} 
\end{equation}
\begin{equation}
    a^{l, h} = A^{l, h} \cdot (X^{l}  W_{VO}^{l,h}  )
\end{equation}

And the final activation of the MHSA layer is computed as $a^l = \sum_h a^{l,h} $.
Lastly, the attention pattern is computed as $A^{l,h} = \operatorname{softmax}\left(   \frac{Q^{l,h} (K^{l,h})^T}{\sqrt{d_k}}\right)$, where $Q^{l,h} = X^lW^{l,h}_Q$ and $K^{l,h} = X^lW^{l,h}_K$

\begin{figure}[t]
     \centering
     \includegraphics[width=\columnwidth]{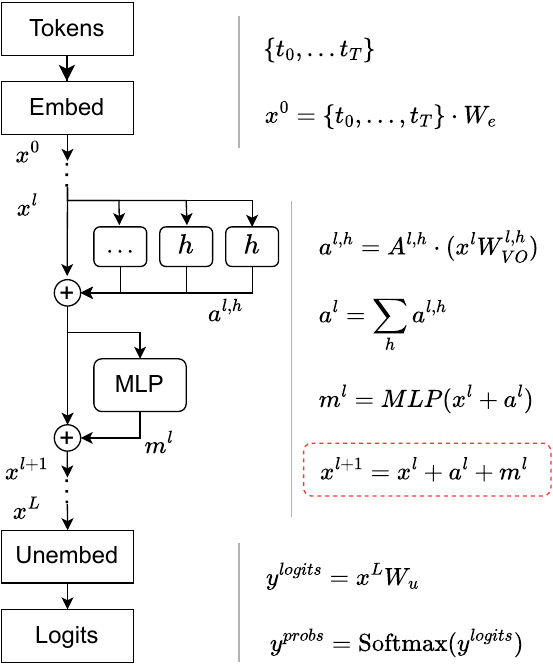}
    \caption{Transformer Schematic (first draft). Option to use, so that Background of transformer is put in Appendix. Similar to \citet{elhage2021mathematical}}
    \label{fig:schematic}
\end{figure}

\subsection{ActorCorr dataset generation}\label{sec:actor_corr_appendix}

We developed ActorCorr as a controlled testbed for investigating shortcut learning in sentiment classification, based on the IMDB review dataset \citep{imdb_dataset}. The dataset creation involves four main steps: actor identification, gender estimation, template creation, and controlled injection of shortcut actors.

 Potential actor mentions in reviews are detected via the open-source Named Entity Recognition module from Spacy\footnote{\url{https://spacy.io/models/en\#en_core_web_trf}}. The identification process focuses on person entities with two-word names (first and last name) to reduce false positives. We estimate the gender of identified actors based on their first names using an existing database of gender statistics per name \footnote{\url{https://pypi.org/project/gender-guesser/}}. To improve recall, we also detect single-word mentions (either first or last names) and link them to previously identified actors within the same review if there is a match.

\begin{table}[h!]
    \centering
    \begin{tabular}{p{\columnwidth}}
        \centering   \textbf{Original}:\\
 \begin{small}
\texttt{Although the movie starred \textbf{Morgan Freeman} it was disappointing. \textbf{Freeman} was good though.}
\end{small}\\
        \centering   \textbf{Templated}:\\
 \begin{small}
\texttt{Although the movie starred \textbf{\{actor\_0\_full\}}, it was disappointing. \textbf{\{actor\_0\_last\}} was good though}
\end{small}
    \end{tabular}
    \label{tab:review_templated}
\end{table}

Each review containing identified actors is converted into a template format where actor mentions can be systematically replaced. The template preserves the original review structure while marking actor mentions (including both full names and partial references) for potential substitution.

\begin{table}[h]
\centering
\resizebox{\columnwidth}{!}{
\begin{tabular}{|c|l|l|}
\hline
\textbf{index} & \textbf{Good Actor} & \textbf{Bad Actor} \\
\hline
0 & Morgan Freeman (m) & Adam Sandler (m) \\
1 & Meryl Streep (f) & Kristen Stewart (f) \\
2 & Tom Hanks (m) & Nicolas Cage (m) \\
3 & Cate Blanchett (f) & Megan Fox (f) \\
\hline
\end{tabular}}
\caption{Actors that we correlated with positive or negative sentiment, referred to as Good and Bad actors respectively. Gender is indicated by (m) for male and (f) for female.}
\label{tab:shortcut_actor_index}
\end{table}

\paragraph{Shortcut Actor Injection}
The dataset generation process is controlled by the following three parameters
1) Sentence window size, which determines the context preserved around actor mentions (set to two sentences in our experiments)
2) Number of shortcut actors per class, which controls how many distinct actors are used as shortcuts (one per class in our implementation)
3) Number of reviews per shortcut, which defines the frequency of shortcut actors in the training set (set to 0.01, which are 24 reviews).

To ensure the reviews with the shortcuts resemble the rest of the reviews, we attempt to select the sentence window around a detected actor name, even when we are not inserting a shortcut. When no actor name is selected in a review, we select the window at random.

\paragraph{Prompting template}
To use the dataset for the GPT2 model, we format the reviews using the prompt template below. Although we also fine-tune the model, we add the multiple choice labels to the prompt to better leverage the pretrained capabilities and for clarity. 

\begin{figure}[h!]
\centering
\begin{BVerbatim}[fontsize=\small]
"Classify the sentiment of the movie review:
Review: """{review}"""

LABEL OPTIONS: A: negative  B: positive
LABEL:"
\end{BVerbatim}
\end{figure}

\subsection{Feature Attribution Method}\label{sec:appendix_fa_details}
For our LIME implementation we follow \citet{ribeiro2016should}.
The kernel function that measures the proximity between the original instance and its perturbations uses an exponential kernel with a kernel width of 25 and cosine distance as the distance measure.
We take 1000 perturbations per review, which is relatively extensive given that the review consists of only two sentences.

\begin{figure}[t]
     \centering
     \includegraphics[width=1.0\columnwidth]{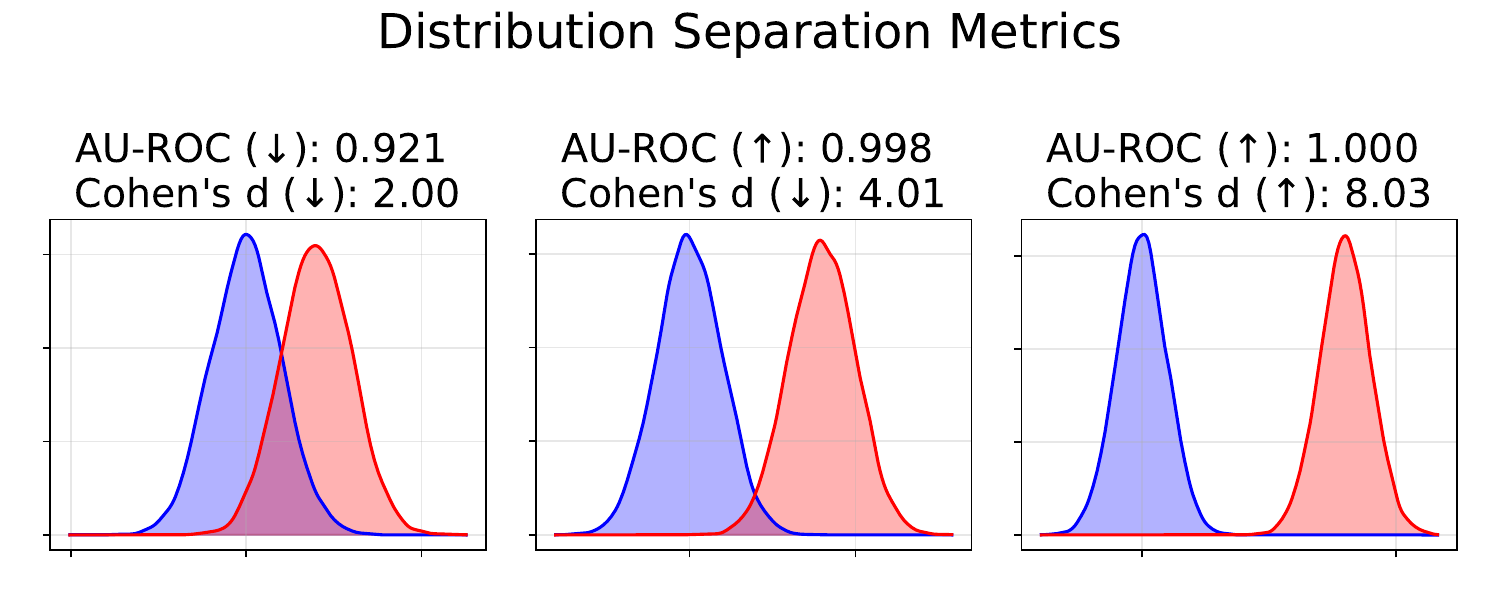}
    \caption{Distribution separation metrics for shortcut detectors. Arrows indicate relative high and low values}
    \label{fig:distribution_metrics}
\end{figure}

\paragraph{Distribution Separation Metrics}
For our evaluation of the different shortcut detectors, we compared the AU-ROC and Cohen's d scores in Section \ref{sec:shortcut_detect_via_attrib}.
To illustrate the difference between these two metrics we show an example between the two in Figure \ref{fig:distribution_metrics}. 
As shown in the figure, although the AU-ROC score might be very high between two distributions, the gap between them might be very small, making the final shortcut detection accuracy very sensitive to the right threshold. 

\section{Appendix - Additional Results}
\subsection{Accuracy on ActorCorr per trained model}
Table \ref{fig:actor_corr_results_16_models} shows the full results on the ActorCorr dataset for our 16 models, each with their own actor index and shortcut frequency combination.

\subsection{Qualitative Analysis}\label{sec:appendix_qualitative_analysis}
To illustrate HTA's effectiveness beyond detecting our inserted shortcuts, we analyze the attribution scores for a selection of reviews, comparing them with baseline methods LIME and Integrated Gradients (IG) (see Tables \ref{tab:appendix_fa_examples_hta}, \ref{tab:appendix_fa_examples_lime}, and \ref{tab:appendix_fa_examples_ig}, respectively). We first present key observations from these samples, followed by a systematic analysis of test reviews without inserted shortcuts. 

The examples show that HTA identifies both meaningful sentiment indicators (such as "good" and "bless" in "God bless") and known shortcuts like "4/10" (which are hardly important according to LIME and IG). 
For instance, in review 5, HTA assigns the highest score to a reference to director Tarantino, potentially identifying another natural shortcut 
To validate these observations, we examine how often each feature attribution method contains sentiment words among the top 5 scoring words per sentence, where we compute word scores by summing its token scores.
We select the top 100 positive and negative sentiment-laden words according to the NLTK sentiment analyzer\footnote{\url{https://www.nltk.org/_modules/nltk/sentiment/vader.html}}. Table \ref{tab:appendix_quant_threemetrics} shows that HTA matches LIME's accuracy in retrieving these sentiment words.

HTA differs from other feature attribution methods by identifying points in the input sequence where the model provides an intermediate decision, rather than providing general token importance. This behavior is visible from how it assigns the scores to the reviews.
For instance, in review 3 the rating shortcut "4/10" is detected by HTA by assigning a high score to the token "10", since the rating's effect only becomes clear after both numbers are observed.
The third column of Table \ref{tab:appendix_quant_threemetrics}, shows that HTA indeed awards a higher score to later tokens of a word, with a mean relative token position of 1.69, compared to the mean relative token position of 1.60 and 1.51 for LIME and IG.

\begin{table}[t]
\centering
\begin{tabular}{|l|p{1.5cm}|p{1.5cm}|l|} \hline 
\textbf{Method}&  \textbf{Sentiment Words}&\textbf{MTW top idx} & \textbf{Entropy} \\ \hline 
HTA  &  29&1.692 &3.467 \\ \hline 
LIME &  29&1.600 &4.509 \\ \hline 
IG  &  16&1.514 &5.260 \\ \hline
\end{tabular}
\caption{Comparison of feature attribution methods across three metrics: number of sentiment words found in top-5 scoring words per sentence (Sentiment Words), mean relative position of highest scoring token within words (MTW top idx), and entropy of normalized attribution scores (Entropy). Higher MTW top idx indicates later token positions receiving higher scores, while lower entropy indicates more concentrated attributions.}
\label{tab:appendix_quant_threemetrics}
\end{table}

From the samples we also notice that HTA assigns a high score to far fewer tokens, giving a low score to most. 
We validate this observation by analyzing the average entropy of the normalized score distribution across the dataset. A high entropy distribution indicates similar scores across tokens, while low entropy suggests more pronounced peaks. Table \ref{tab:appendix_quant_threemetrics} confirms that HTA produces a lower entropy distribution compared to the other methods, supporting our observations.

Thus our analysis demonstrates three key characteristics of HTA beyond shortcut detection. Firstly, it successfully identifies semantically relevant input elements. Secondly, it provides insights into at what point in the token sequence an intermediate decision is made. Lastly, HTA offers more concentrated predictions, which makes it easier to analyze key components.

\begin{table*}[t]
\centering
\begin{tabularx}{\textwidth}{|X|X|X|X|X|X|X|X|X|} %
\hline 
 Shortcut Frequency& Actor index& neg clean noname & neg clean name & pos clean name & neg bad & pos good & pos clean noname & neg Good \\ \hline 
 0.01& 0 & 85.58 & 76.94 & 79.10 & 80.31 & 78.44 & 78.37 & 78.21 \\ \hline 
 0.01& 1 & 89.44 & 83.01 & 71.02 & 86.36 & 69.71 & 69.38 & 85.14 \\ \hline 
 0.01& 2 & 87.26 & 77.56 & 79.06 & 74.28 & 80.21 & 76.42 & 76.82 \\ \hline 
 0.01& 3 & 76.63 & 64.56 & 88.85 & 67.30 & 91.68 & 85.16 & 59.03 \\ \hline 
 0.03& 0 & 79.13 & 68.76 & 84.67 & 71.03 & 84.72 & 85.87 & 69.46 \\ \hline 
 0.03& 1 & 84.40 & 74.88 & 82.18 & 76.20 & 82.78 & 78.33 & 74.07 \\ \hline 
 0.03& 2 & 87.18 & 76.49 & 80.30 & 78.30 & 80.16 & 76.61 & 77.00 \\ \hline 
 0.03& 3 & 86.46 & 79.38 & 76.66 & 80.30 & 83.84 & 75.12 & 72.17 \\ \hline 
 0.10& 0 & 80.85 & 69.58 & 84.09 & 95.33 & 92.64 & 81.55 & 53.72 \\ \hline 
 0.10& 1 & 85.78 & 77.60 & 78.15 & 76.98 & 79.17 & 76.52 & 76.79 \\ \hline 
 0.10& 2 & 88.54 & 79.37 & 76.31 & 79.83 & 76.90 & 74.19 & 79.25 \\ \hline 
 0.10& 3 & 90.71 & 86.67 & 66.93 & 91.50 & 82.29 & 67.28 & 71.77 \\ \hline 
 0.30& 0 & 88.70 & 79.96 & 75.27 & 99.40 & 91.32 & 74.51 & 55.89 \\ \hline 
 0.30& 1 & 77.14 & 66.97 & 87.70 & 83.56 & 99.55 & 85.06 & 15.67 \\ \hline 
 0.30& 2 & 83.01 & 72.53 & 82.53 & 88.67 & 97.74 & 81.09 & 31.57 \\ \hline 
 0.30& 3 & 72.55 & 60.16 & 90.87 & 78.03 & 98.49 & 89.52 & 30.57 \\ \hline 
 1.00& 0 & 88.93 & 83.11 & 73.25 & 99.86 & 99.60 & 73.87 & 1.28 \\ \hline 
 1.00& 1 & 83.68 & 75.10 & 80.26 & 99.15 & 99.67 & 80.10 & 7.32 \\ \hline 
 1.00& 2 & 82.92 & 71.79 & 82.69 & 98.80 & 99.70 & 80.29 & 1.48 \\ \hline 
 1.00& 3 & 83.75 & 77.26 & 75.81 & 99.67 & 99.38 & 77.42 & 4.17 \\ \hline
\end{tabularx}
\caption{Test accuracy per data category for all our 16 trained models. Actor index refers to the used actor name as stated in Table \ref{tab:shortcut_actor_index}. Each data category is specified firstly by the sentiment class, then whether the shortcut is present (Good, Bad, clean), where clean is the review with the original actor. Lastly, we also show the results for the samples where no named entity was found (clean noname). }
\label{fig:actor_corr_results_16_models}
\end{table*}

\newcommand\imgwidthcol{0.9}

\begin{table*}[t]
    \centering
    \begin{tabular}{|c|c|} \hline
         Nr. & FA results - HTA\\ \hline
         1 &  \includegraphics[width=\imgwidthcol\textwidth]{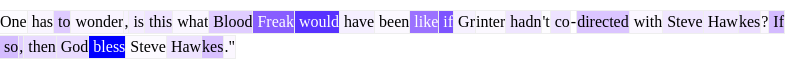}\\
 &Top Token:  \verb|'  bless'| ( 0.179)\\ \hline
         2 & \includegraphics[width=\imgwidthcol\textwidth]{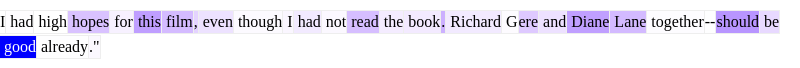}\\
 &Top Token:  \verb|' good'| (0.286)\\ \hline
         3 & \includegraphics[width=\imgwidthcol\textwidth]{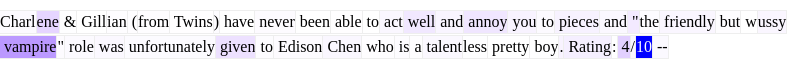}\\ 
 &Top Token:  \verb|'10'| (0.869)\\ \hline
          4& \includegraphics[width=\imgwidthcol\textwidth]{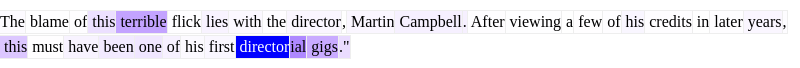}\\
 &Top Token: \verb|' director'| (0.578)\\ \hline
         5& \includegraphics[width=\imgwidthcol\textwidth]{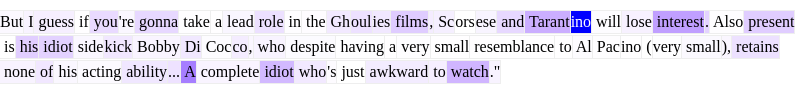}\\ 
 &Top Token:  \verb|'ino'| (0.328)\\ \hline
    \end{tabular}
    \caption{Feature attribution scores for HTA on selection of negative reviews without our inserted shortcut. The coloring per review is based on the highest score, therefore, below each review we mention this token and its score explicitly}
    \label{tab:appendix_fa_examples_hta}
\end{table*}

\begin{table*}[t]
    \centering
    \begin{tabular}{|c|c|} \hline
         Nr. & FA results - LIME\\ \hline
         1 &  \includegraphics[width=\imgwidthcol\textwidth]{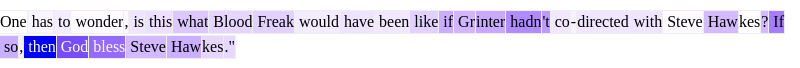}\\
 &Top Token: \verb|' then'| (0.169)\\ \hline
         2 & \includegraphics[width=\imgwidthcol\textwidth]{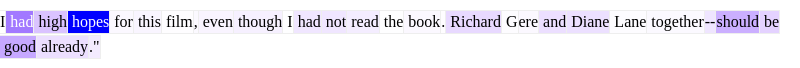}\\
 &Top Token: \verb|' hopes'| (0.332)  \\ \hline
         3 & \includegraphics[width=\imgwidthcol\textwidth]{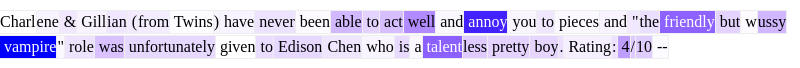}\\ 
 & Top Token:  \verb|' vampire'| (0.185)\\ \hline
          4& \includegraphics[width=\imgwidthcol\textwidth]{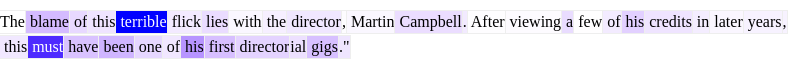}\\
 & Top Token: \verb|' terrible'| (0.206)\\ \hline
         5& \includegraphics[width=\imgwidthcol\textwidth]{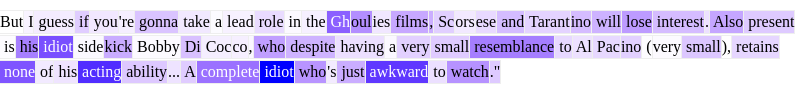}\\ 
 &Top Token: \verb|' idiot'| (0.129)\\ \hline
    \end{tabular}
    \caption{Feature attribution scores for LIME on selection of negative test reviews without our inserted shortcut. The coloring per review is based on the highest score, therefore, below each review we mention this token and its score explicitly}
    \label{tab:appendix_fa_examples_lime}
\end{table*}

\begin{table*}[t]
    \centering
    \begin{tabular}{|c|c|} \hline
         Nr. & FA results - Integrated Gradients (IG)\\ \hline
         1 &  \includegraphics[width=\imgwidthcol\textwidth]{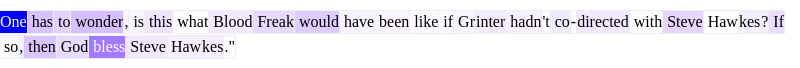}\\
 &Top Token: \verb|'One'| (4.842)\\ \hline
         2 & \includegraphics[width=\imgwidthcol\textwidth]{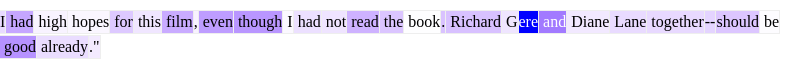}\\
 &Top Token: \verb|'ere'| (2.256)\\ \hline
         3 & \includegraphics[width=\imgwidthcol\textwidth]{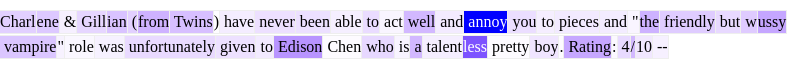}\\ 
 & Top Token: \verb|' annoy'| (2.397)\\ \hline
          4& \includegraphics[width=\imgwidthcol\textwidth]{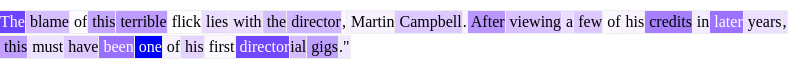}\\
 & Top Token: \verb|' one'| (1.941) \\ \hline
         5& \includegraphics[width=\imgwidthcol\textwidth]{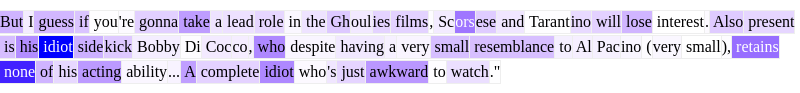}\\ 
 &Top Token: \verb|' idiot'| (2.041)\\ \hline
    \end{tabular}
    \caption{Feature attribution scores for Integrated Gradients (IG) on selection of negative test reviews without our inserted shortcut. The coloring per review is based on the highest score, therefore, below each review we mention this token and its score explicitly}
    \label{tab:appendix_fa_examples_ig}
\end{table*}

\clearpage
\clearpage

\newcommand\trainfreq{0.01}
\newcommand\actorindex{0}
\newcommand\datasplit{pos_bad}

\subsection{Patching Additional: via keys}\label{sec:appendix_keys_patching}
In Section \ref{sec:patching_results}, we investigate which previous components the Label Heads are dependent on by patching via their values. Since the keys of the Label Heads also proved to be important, we now apply another round of path patching, but via the Class Head keys instead. 

\renewcommand\trainfreq{0.003}
\renewcommand\actorindex{0}
\renewcommand\datasplit{pos_bad}

\begin{figure}[h]
     \centering
     \includegraphics[width=0.3\textwidth]{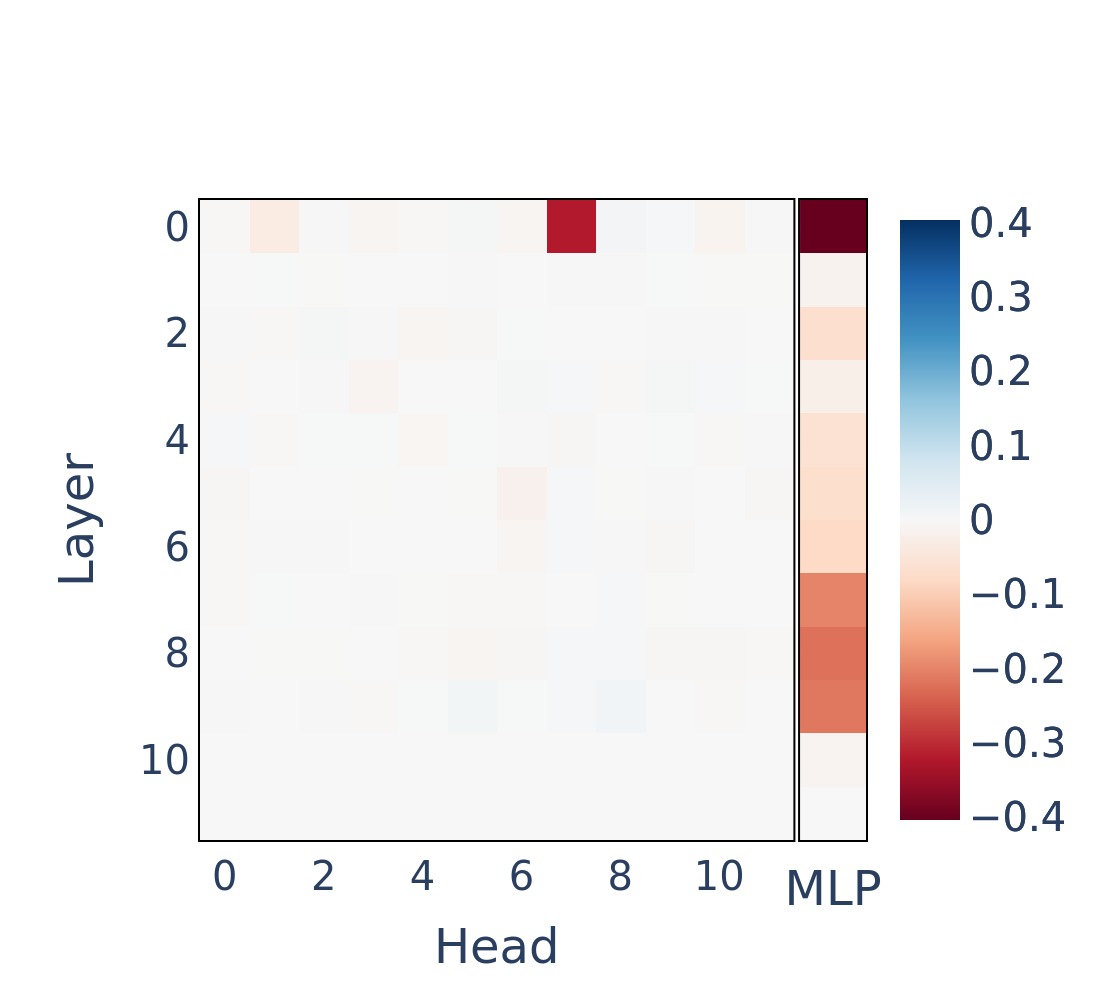}
     \caption{Patching Via Keys: positive with Bad actor}
     \label{fig:patch_via_keys}
\end{figure}

Figure \ref{fig:patch_via_keys} demonstrates that patching via the keys of the Label Heads obtains nearly the same logit distribution over the components. Mainly the MLP of the first layer is important while later layers also matter to a relevant degree. Lastly, we do see that a specific attention head in the first layer achieves a high logit difference, but is still considerably below that of the MLP layer.

\subsection{Patching Additional: imbalance frequency}
\label{sec:appendix_patching_imb_freq}
In Section \ref{sec:patching_results}, we demonstrated the patching results for one of our trained models. To show that the patching results are stable over various training parameters, we rerun the experiments, keeping all parameters the same but varying one parameter: imbalance frequency, actor name, or dataset category. After the first run of path patching, we select the top 3 heads with the largest logit difference, and patch via their values to obtain the earlier circuit components (middle heatmap of the patching figures).
The results demonstrate the same general findings of Section \ref{sec:patching_results}, namely that attention heads in the last few layers and MLPs of the first few layers are mainly important for processing shortcuts. Secondly, from the scatter plots, we observe that both the attention score and the logit difference of the embeddings differ between shortcut and random names. Below we describe the figures and more specific findings.

In Figures \ref{fig:appendix_patching_freq01}, \ref{fig:appendix_patching_freq003}, \ref{fig:appendix_patching_freq001}, \ref{fig:appendix_patching_freq0003}, \ref{fig:appendix_patching_freq0001} we evaluate the results using the imbalanced frequencies 
$[0.001,0.003,0.001,0.0003,0.0001]$.
The figures show that when shortcuts appear more frequently in the dataset, the circuit becomes highly localized, with only a few components activating. Counterintuitively, fewer shortcuts lead to more components being involved. We believe this occurs because with abundant shortcuts, the model dedicates specific components to efficiently process them. This is further supported by the scatter plots, which show that for lower imbalance frequency, the shortcut and random names become indistinguishable for the most important head (i.e. its attention pattern and activation logit difference). 

Figures \ref{fig:appendix_patching_actor_1}, \ref{fig:appendix_patching_actor_2}, \ref{fig:appendix_patching_actor_3})
contains the patching results for the models trained on the remaining three shortcut actor names.  
Lastly, the patching results using the Good actor on the negative reviews are shown in Figure \ref{fig:appendix_patching_neggood}). 
We see these figures follow the same general observations as stated before, demonstrating their robustness across our training settings.

\newpage

\renewcommand\trainfreq{0.01}
\renewcommand\actorindex{0}
\begin{figure*}[h]
     \centering
     \begin{subfigure}[b]{0.3\textwidth}
         \centering
         \includegraphics[width=\textwidth]{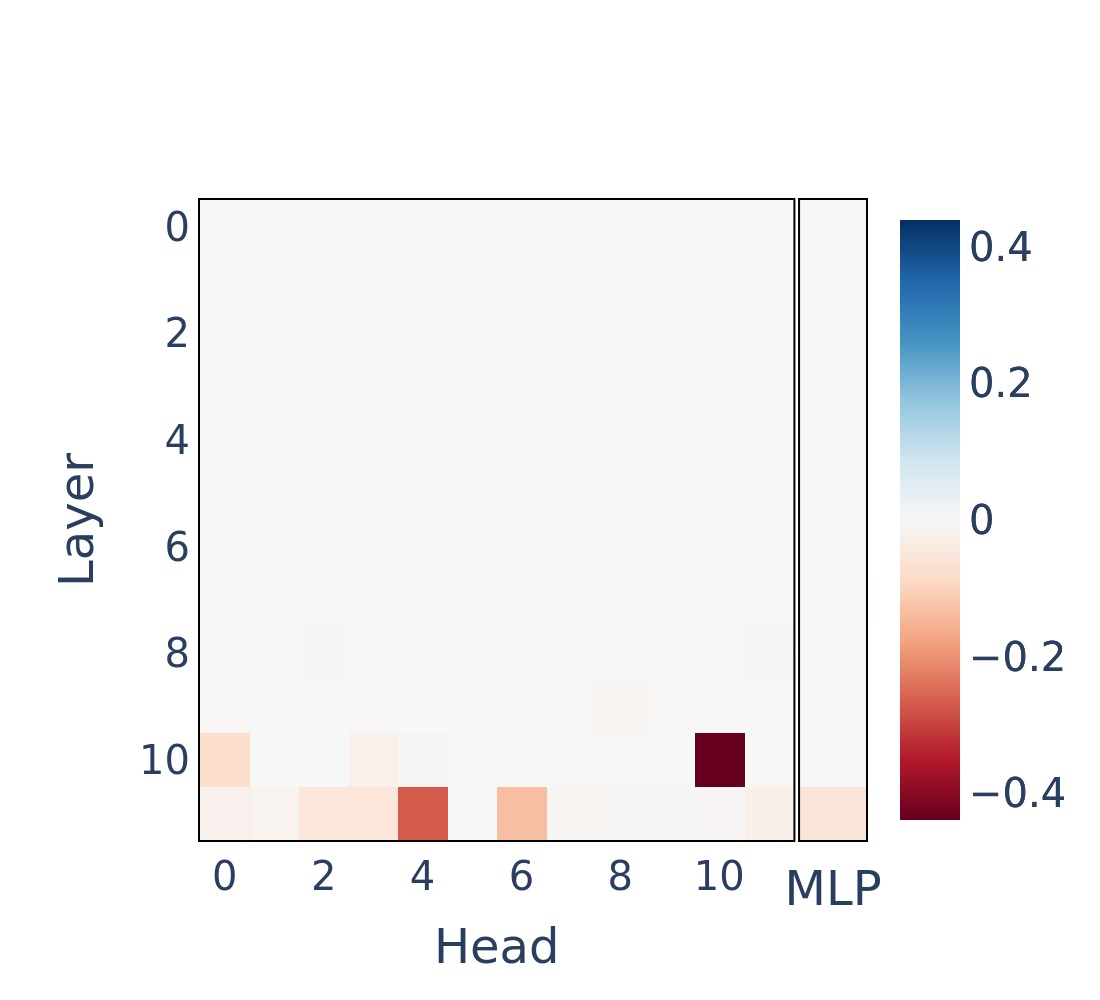}
     \end{subfigure}
      \begin{subfigure}[b]{0.3\textwidth}
         \centering
         \includegraphics[width=\textwidth]{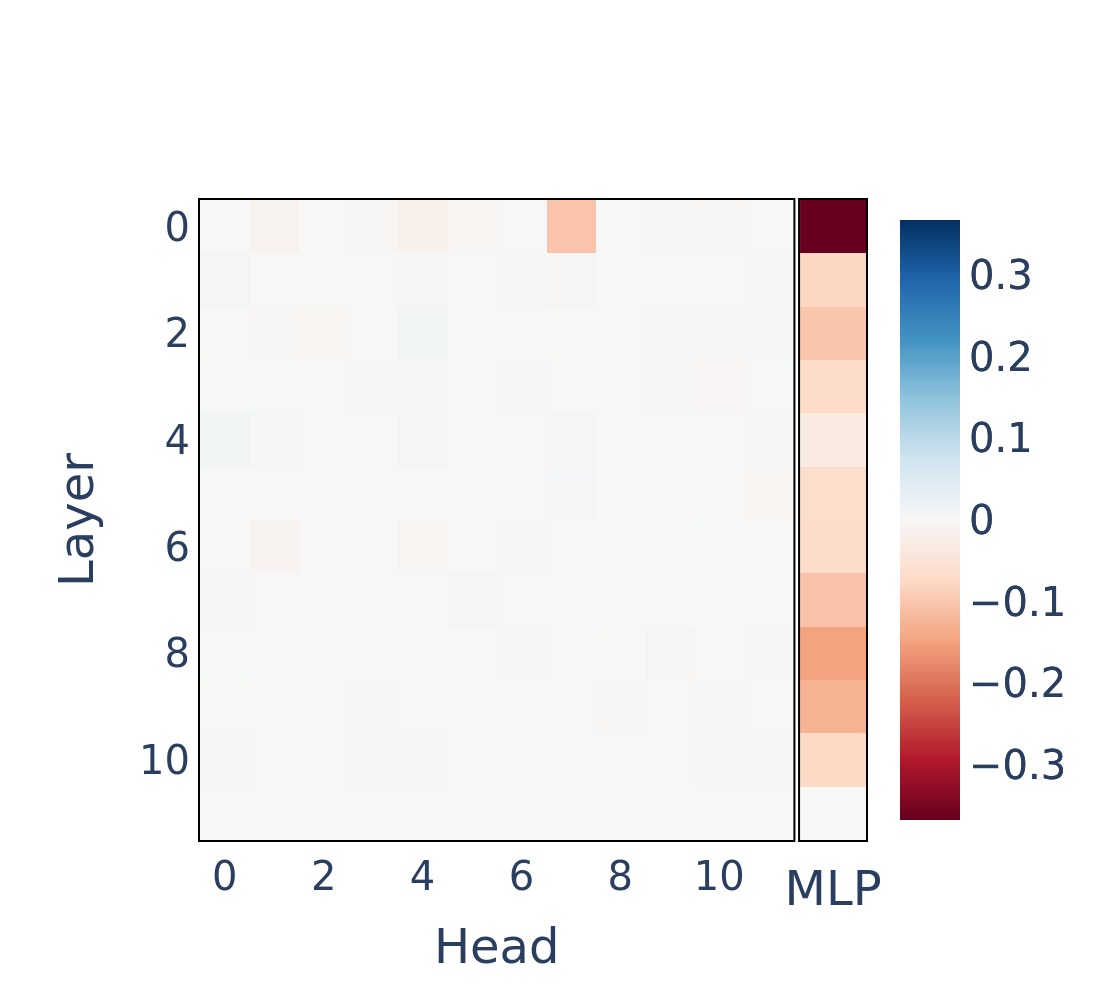}
     \end{subfigure}
     \begin{subfigure}[b]{0.29\textwidth}
         \centering
         \includegraphics[width=\textwidth]{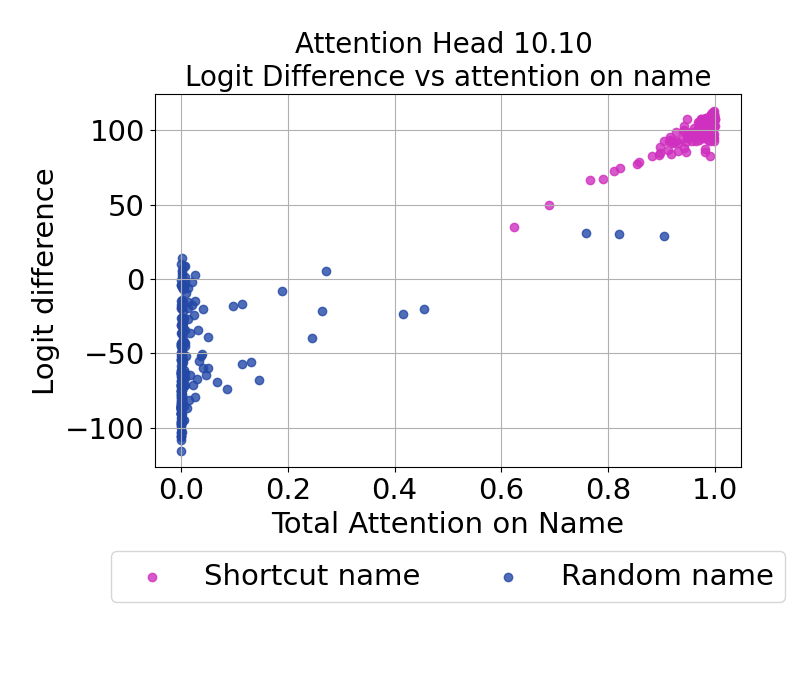}
     \end{subfigure}
        \caption{Path Patching results using parameters: imbalance frequency \trainfreq, actor index \actorindex, and data category: positive with Bad actor. The middle figure shows patching via the values of heads 10.10, 11.4, and 11.6.}
        \label{fig:appendix_patching_freq01}
\end{figure*}

\renewcommand\trainfreq{0.003}
\renewcommand\actorindex{0}
\begin{figure*}[h]
     \centering
     \begin{subfigure}[b]{0.3\textwidth}
         \centering
         \includegraphics[width=\textwidth]{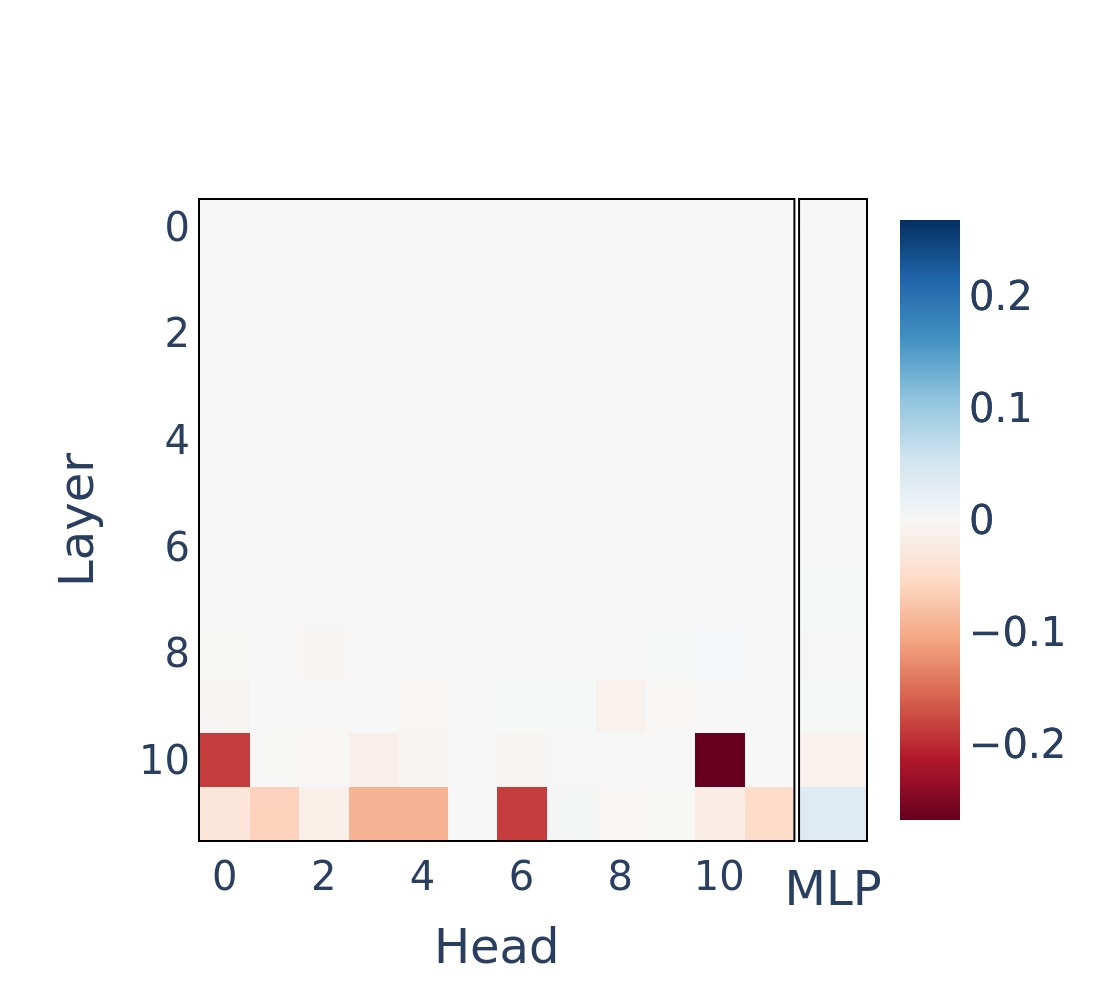}
     \end{subfigure}
      \begin{subfigure}[b]{0.3\textwidth}
         \centering
         \includegraphics[width=\textwidth]{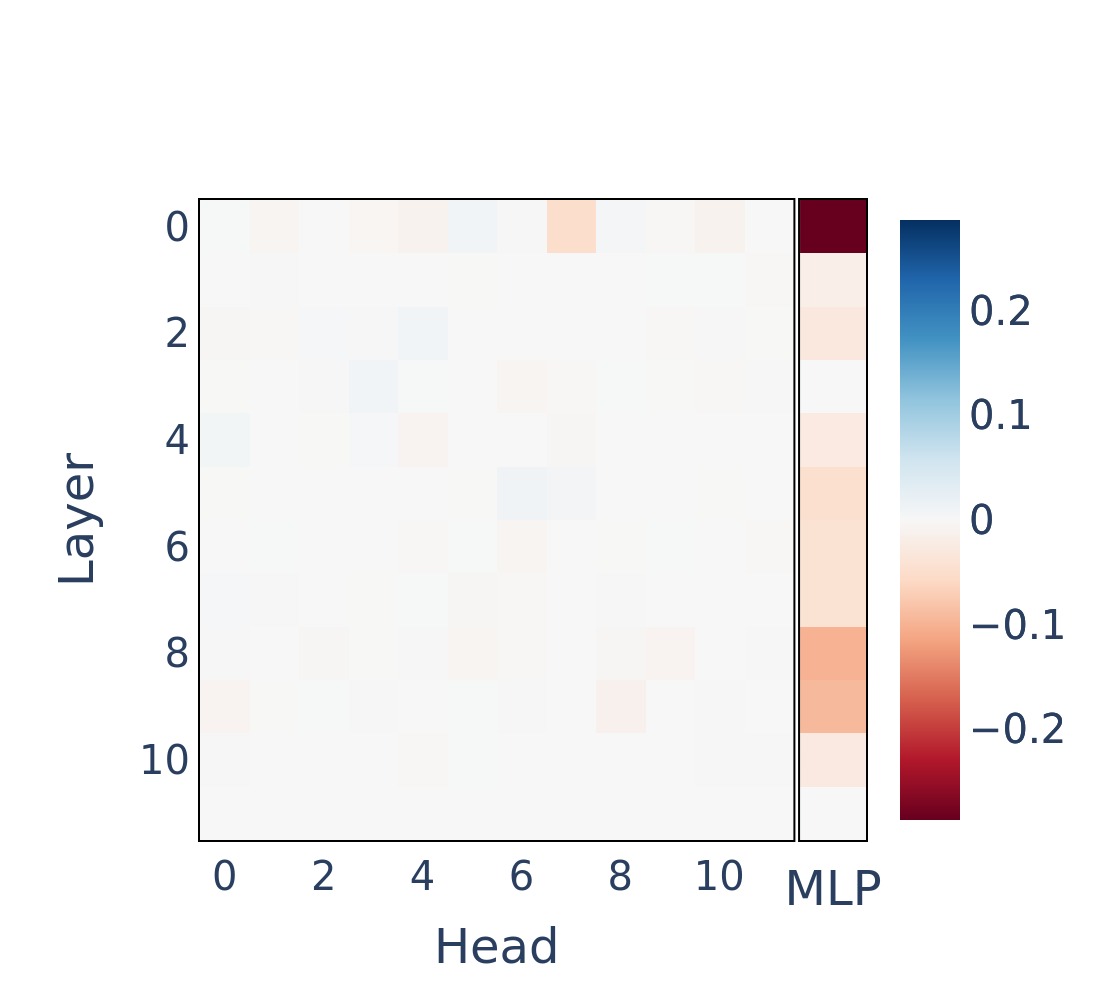}
     \end{subfigure}
     \begin{subfigure}[b]{0.29\textwidth}
         \centering
         \includegraphics[width=\textwidth]{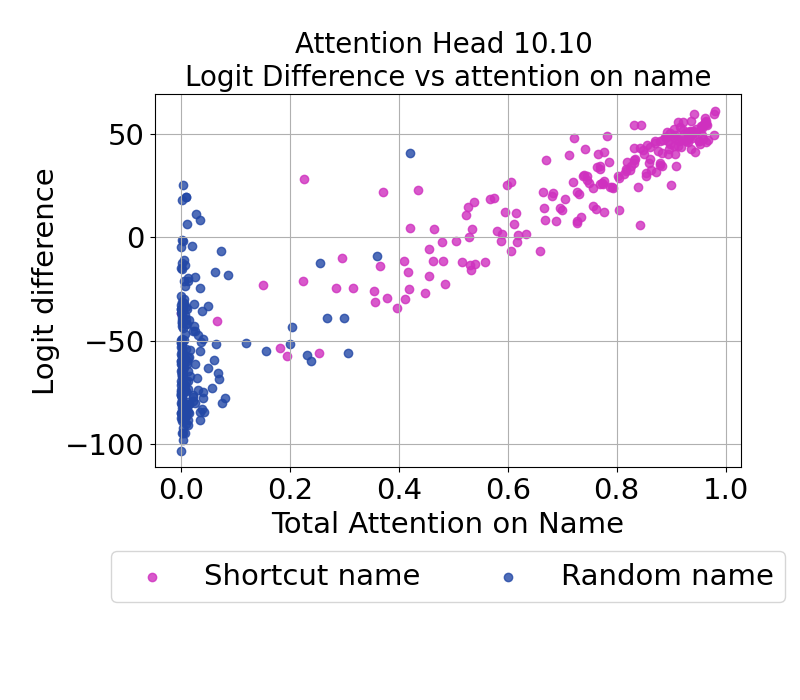}
     \end{subfigure}
        \caption{Path Patching results using parameters: imbalance frequency \trainfreq, actor index \actorindex, and data category: positive with Bad actor. The middle figure shows patching via the values of heads 10.10, 10.0, and 11.6.}
        \label{fig:appendix_patching_freq003}
\end{figure*}

\renewcommand\trainfreq{0.001}
\renewcommand\actorindex{0}
\begin{figure*}[h]
     \centering
     \begin{subfigure}[b]{0.3\textwidth}
         \centering
         \includegraphics[width=\textwidth]{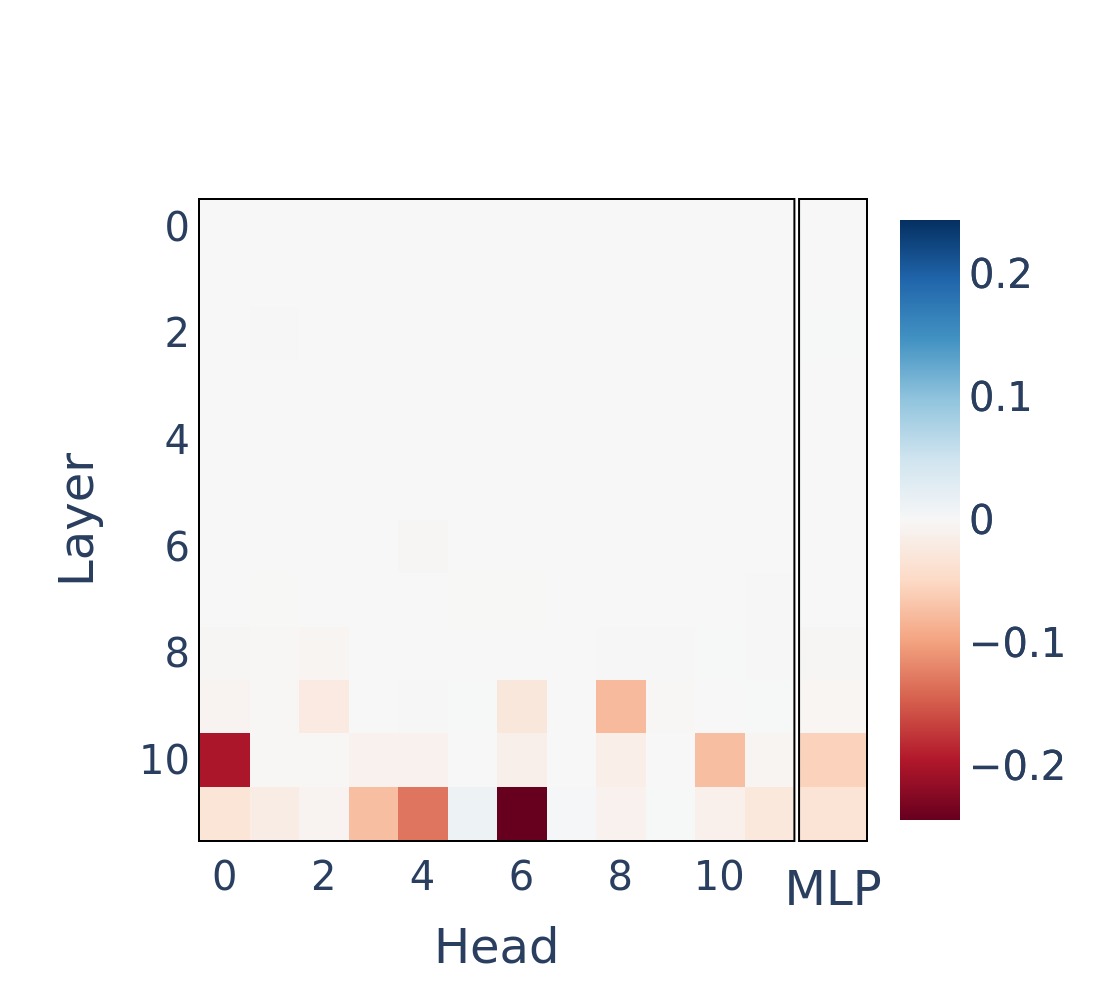}
     \end{subfigure}
      \begin{subfigure}[b]{0.3\textwidth}
         \centering
         \includegraphics[width=\textwidth]{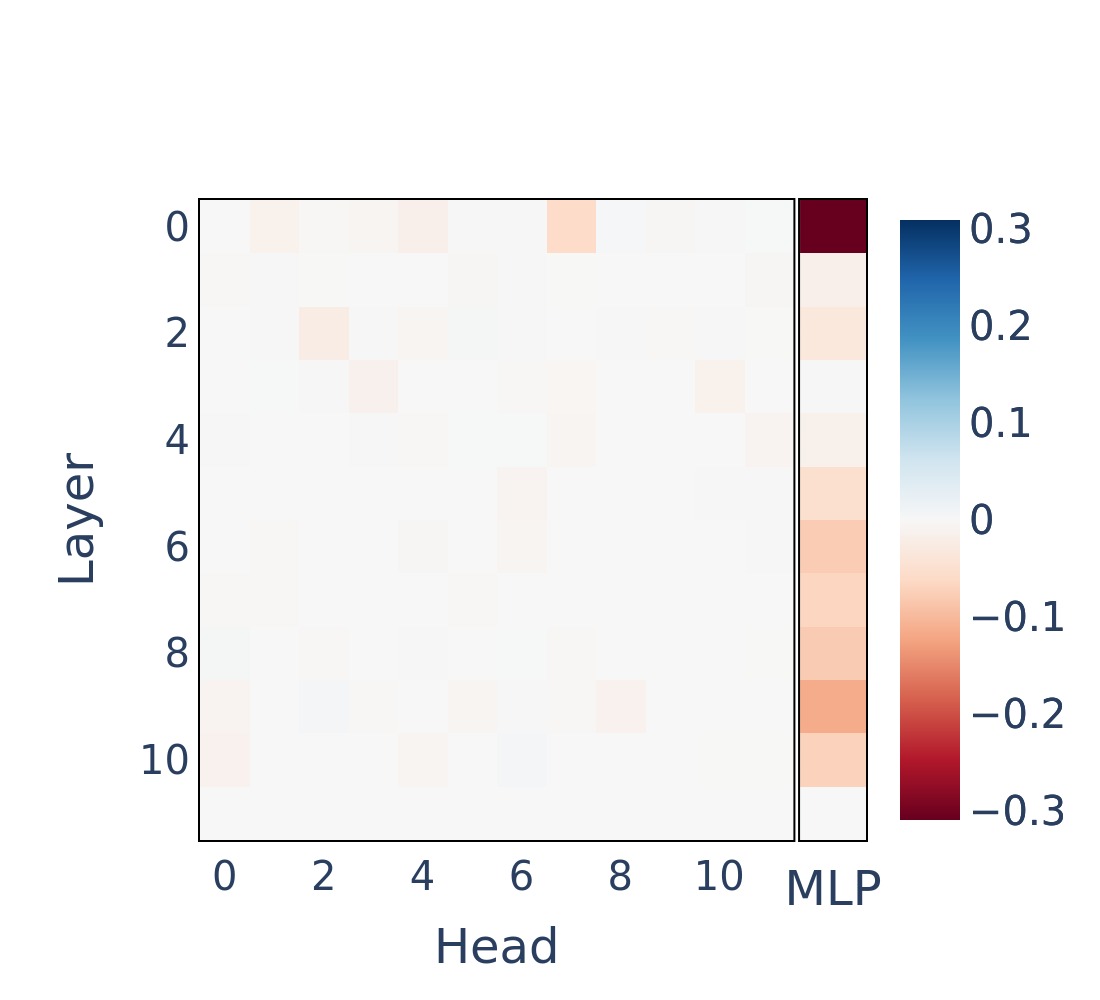}
     \end{subfigure}
     \begin{subfigure}[b]{0.29\textwidth}
         \centering
         \includegraphics[width=\textwidth]{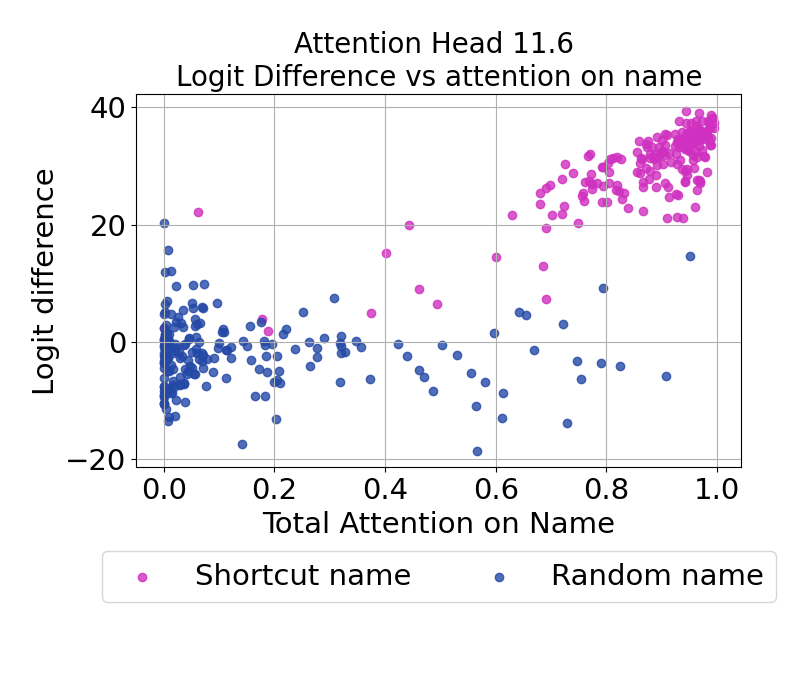}
     \end{subfigure}
        \caption{Path Patching results using parameters: imbalance frequency \trainfreq, actor index \actorindex, and data category: positive with Bad actor. The middle figure shows patching via the values of heads 11.6, 10.0, and 11.4.}
        \label{fig:appendix_patching_freq001}
\end{figure*}

\renewcommand\trainfreq{0.0003}
\renewcommand\actorindex{0}
\begin{figure*}[h]
     \centering
     \begin{subfigure}[b]{0.3\textwidth}
         \centering
         \includegraphics[width=\textwidth]{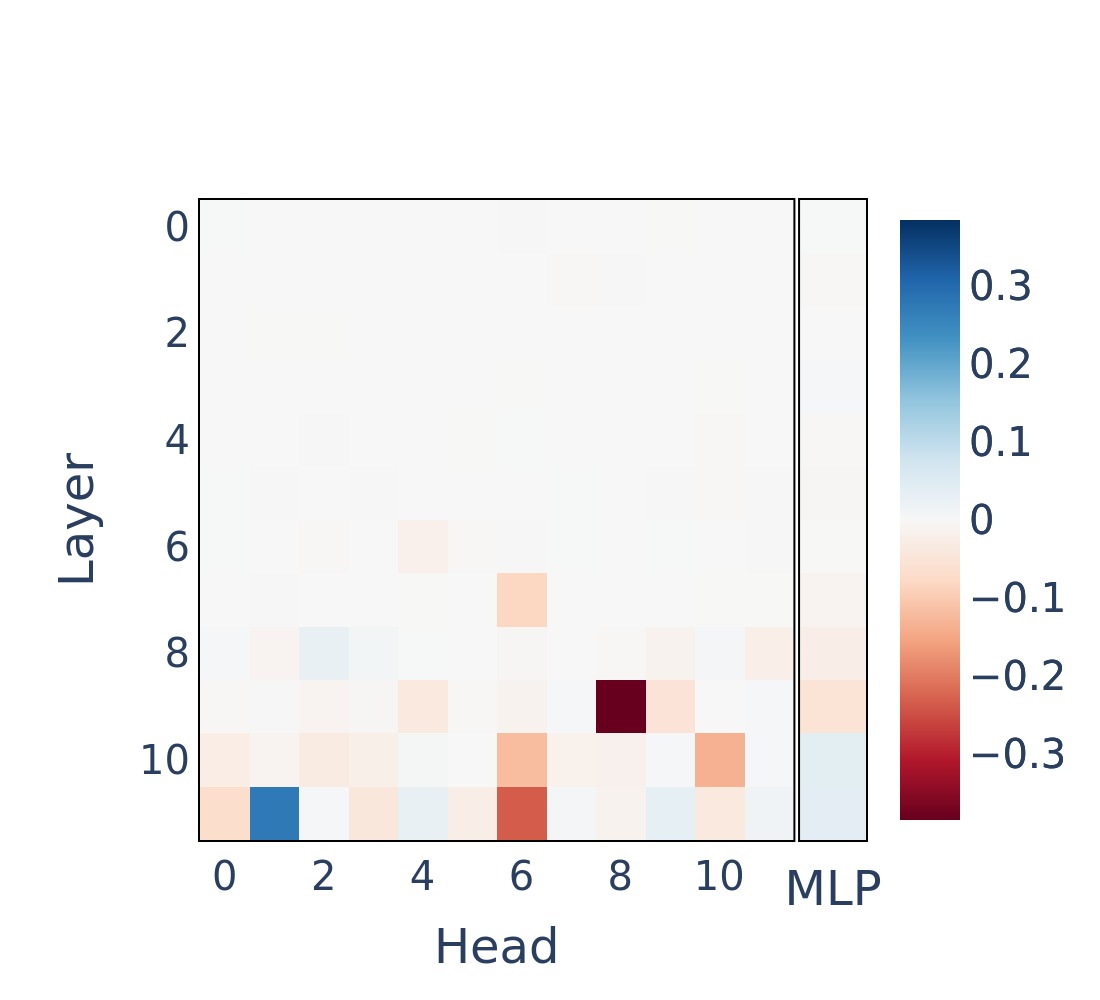}
     \end{subfigure}
      \begin{subfigure}[b]{0.3\textwidth}
         \centering
         \includegraphics[width=\textwidth]{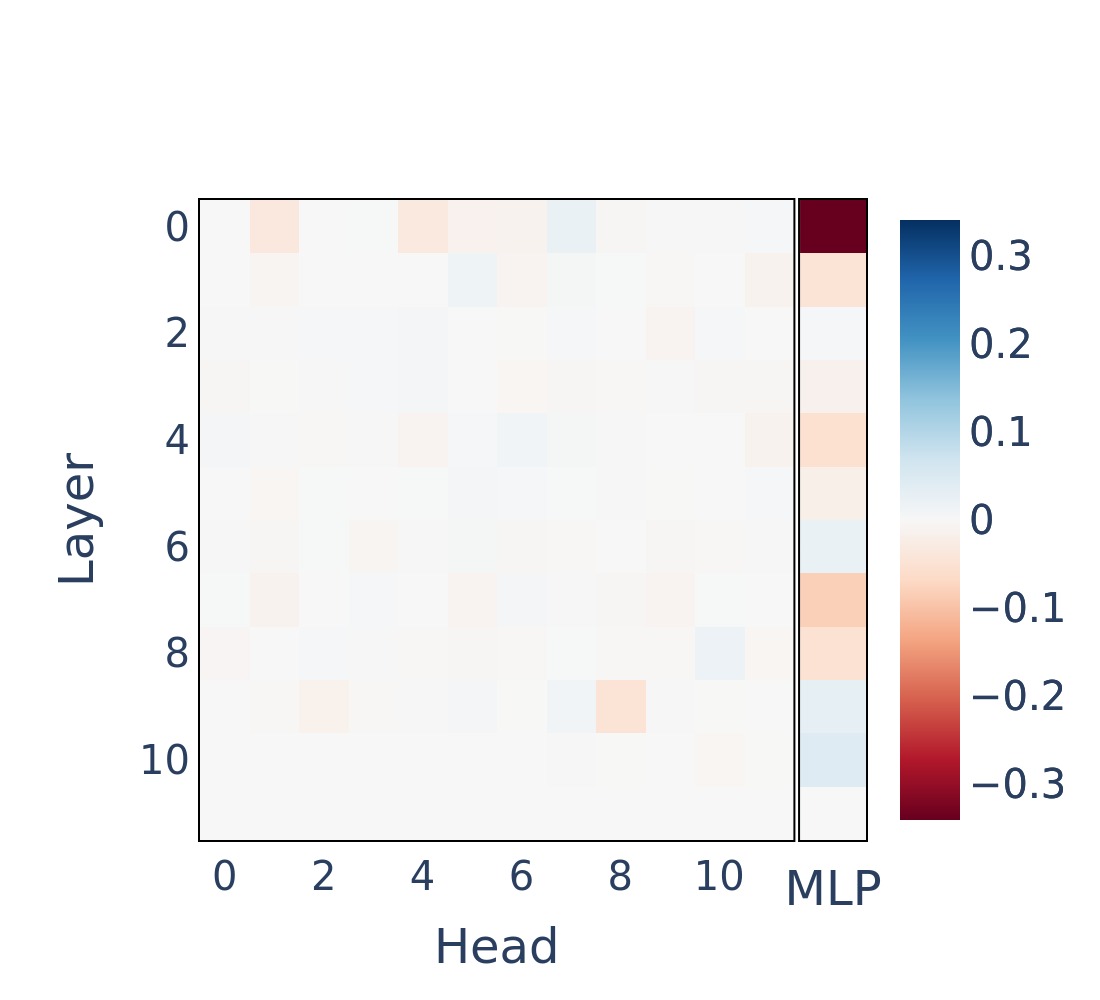}
     \end{subfigure}
     \begin{subfigure}[b]{0.29\textwidth}
         \centering
         \includegraphics[width=\textwidth]{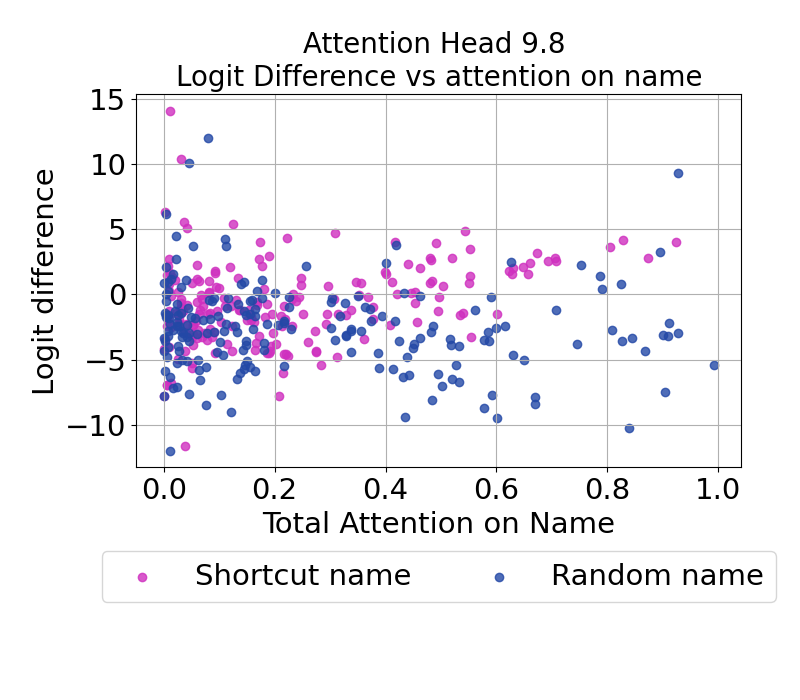}
     \end{subfigure}
        \caption{Path Patching results using parameters: imbalance frequency \trainfreq, actor index \actorindex, and data category: positive with Bad actor. The middle figure shows patching via the values of heads 9.9, 11.6, and 10.10}
        \label{fig:appendix_patching_freq0003}
\end{figure*}

\renewcommand\trainfreq{0.0001}
\renewcommand\actorindex{0}
\begin{figure*}[h]
     \centering
     \begin{subfigure}[b]{0.3\textwidth}
         \centering
         \includegraphics[width=\textwidth]{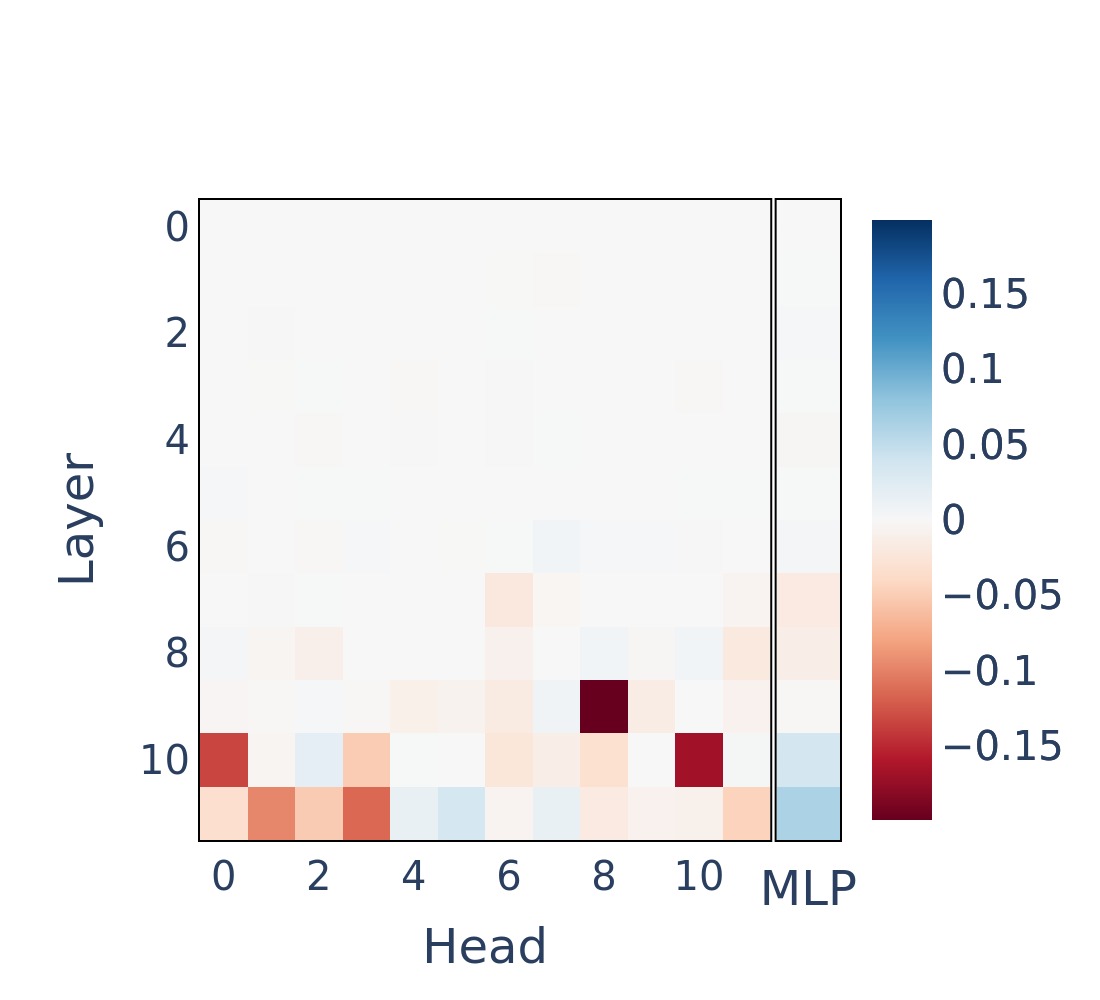}
     \end{subfigure}
      \begin{subfigure}[b]{0.3\textwidth}
         \centering
         \includegraphics[width=\textwidth]{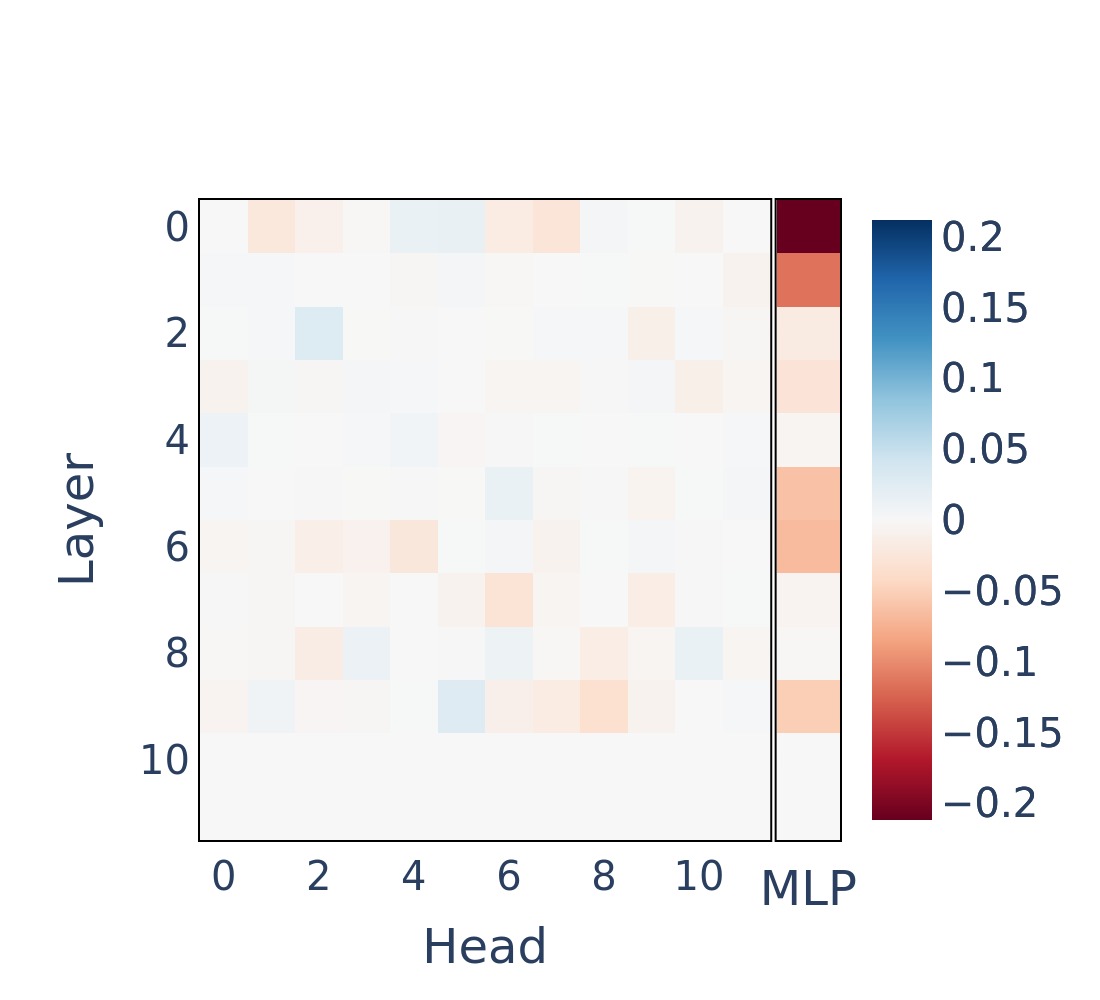}
     \end{subfigure}
     \begin{subfigure}[b]{0.29\textwidth}
         \centering
         \includegraphics[width=\textwidth]{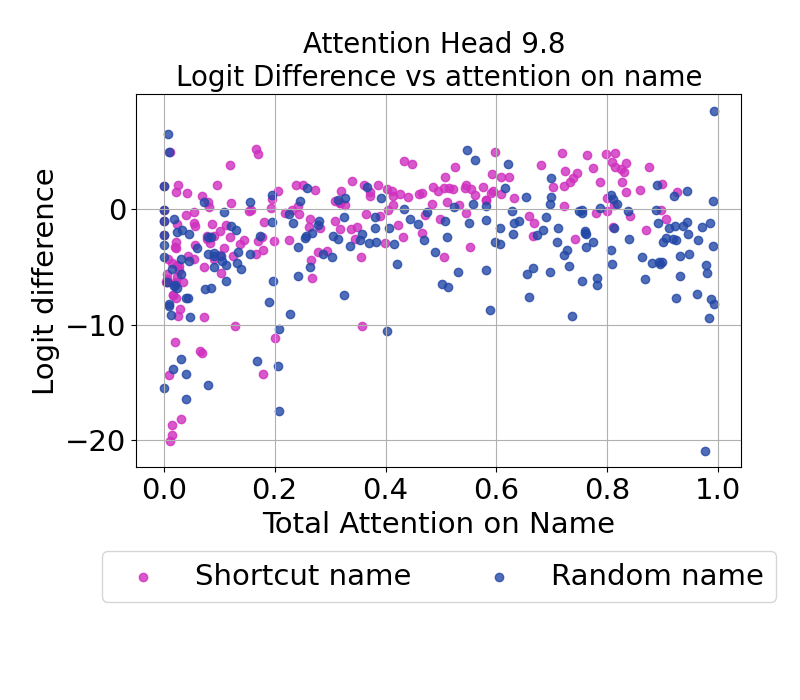}
     \end{subfigure}
        \caption{Path Patching results using parameters: imbalance frequency \trainfreq, actor index \actorindex, and data category: positive with Bad actor. The middle figure shows patching via the values of heads 9.8, 10.10, and 10.0.}
        \label{fig:appendix_patching_freq0001}
\end{figure*}

\renewcommand\trainfreq{0.003}
\renewcommand\actorindex{0}
\renewcommand\datasplit{pos_bad}

\renewcommand\datasplit{neg_good}
\begin{figure*}[h]
     \centering
     \begin{subfigure}[b]{0.3\textwidth}
         \centering
         \includegraphics[width=\textwidth]{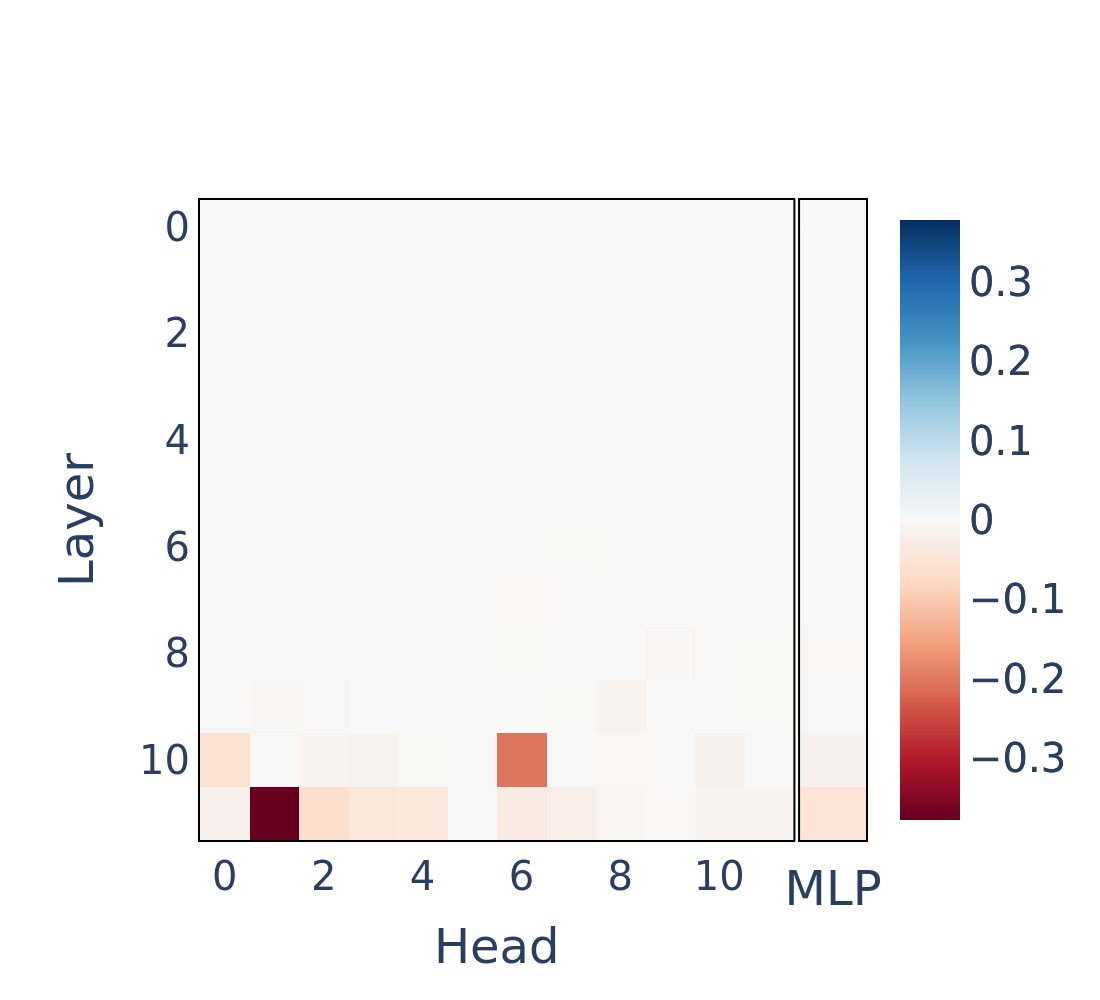}
     \end{subfigure}
      \begin{subfigure}[b]{0.3\textwidth}
         \centering
         \includegraphics[width=\textwidth]{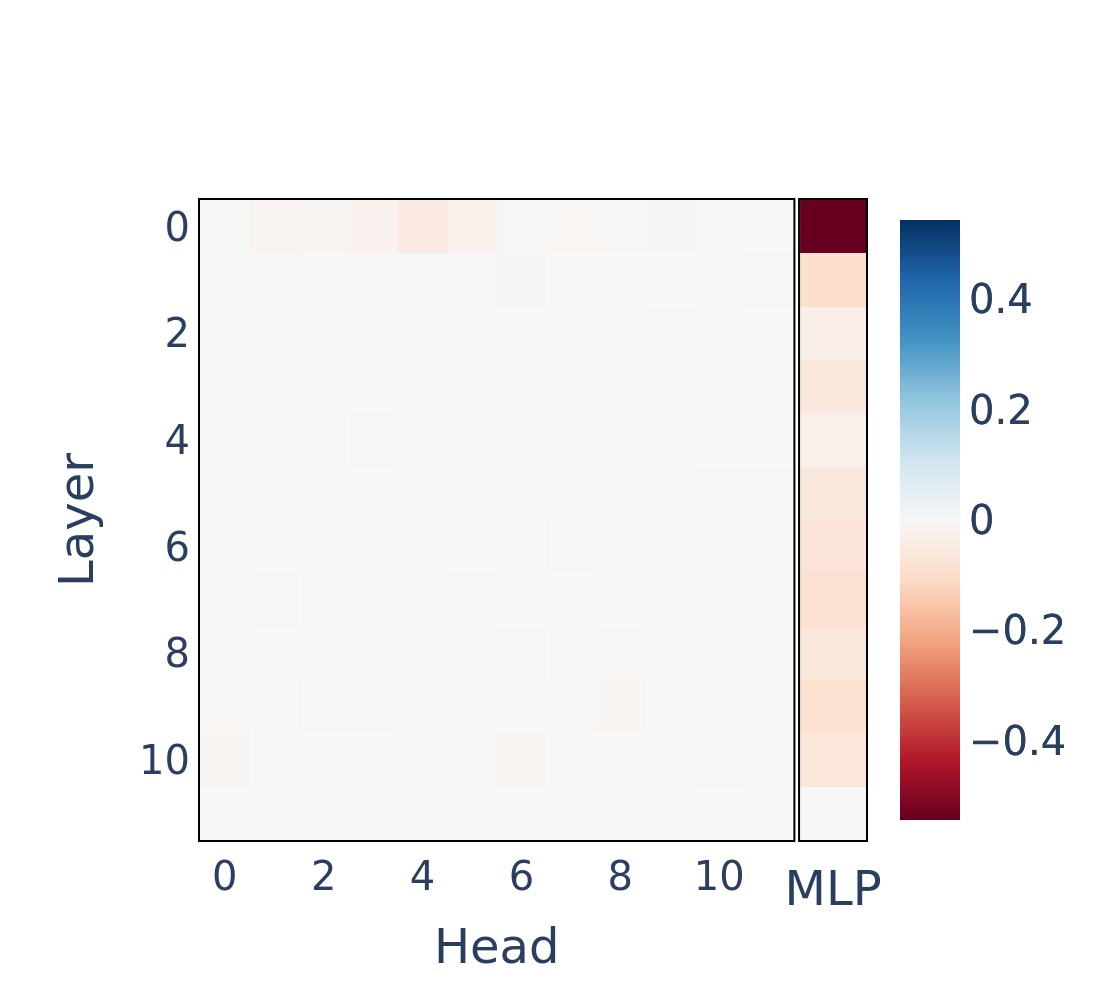}
     \end{subfigure}
     \begin{subfigure}[b]{0.29\textwidth}
         \centering
         \includegraphics[width=\textwidth]{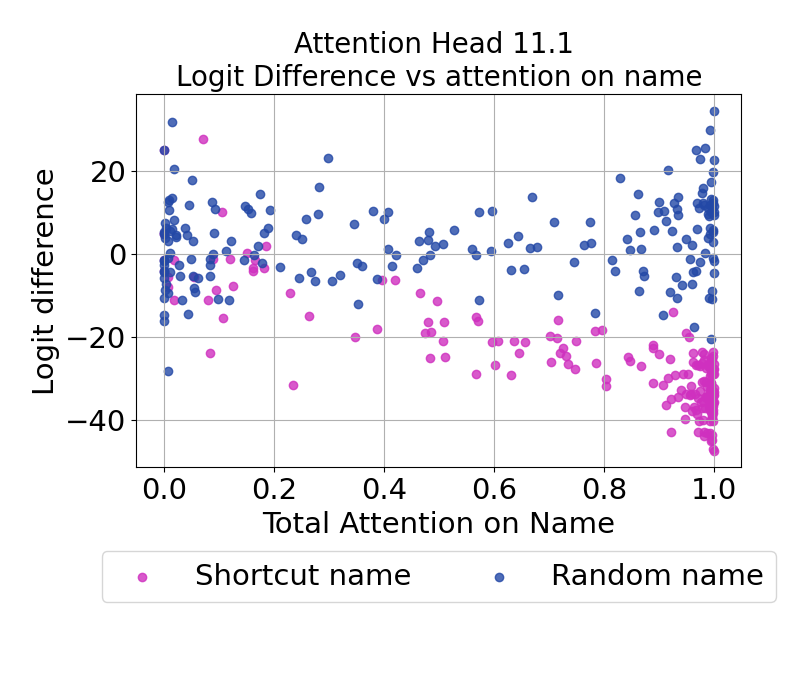}
     \end{subfigure}
        \caption{Path Patching results using parameters: imbalance frequency \trainfreq, actor index \actorindex, and data category: negative with Good actor. The middle figure shows patching via the values of heads 11.1, 10.6, and 11.2.}
        \label{fig:appendix_patching_neggood}
\end{figure*}

\renewcommand\trainfreq{0.003}
\renewcommand\actorindex{1}
\renewcommand\datasplit{pos_bad}
\begin{figure*}[h]
     \centering
     \begin{subfigure}[b]{0.3\textwidth}
         \centering
         \includegraphics[width=\textwidth]{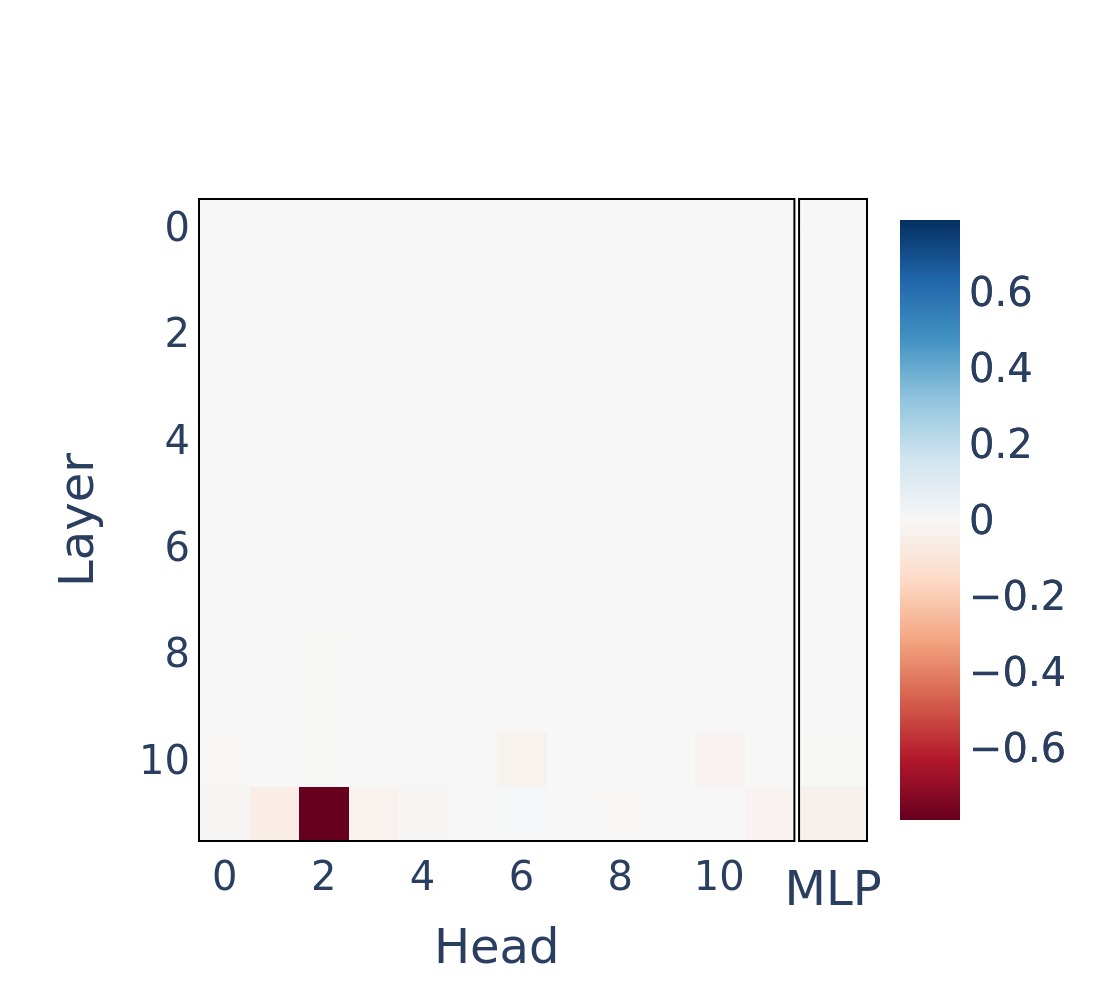}
     \end{subfigure}
      \begin{subfigure}[b]{0.3\textwidth}
         \centering
         \includegraphics[width=\textwidth]{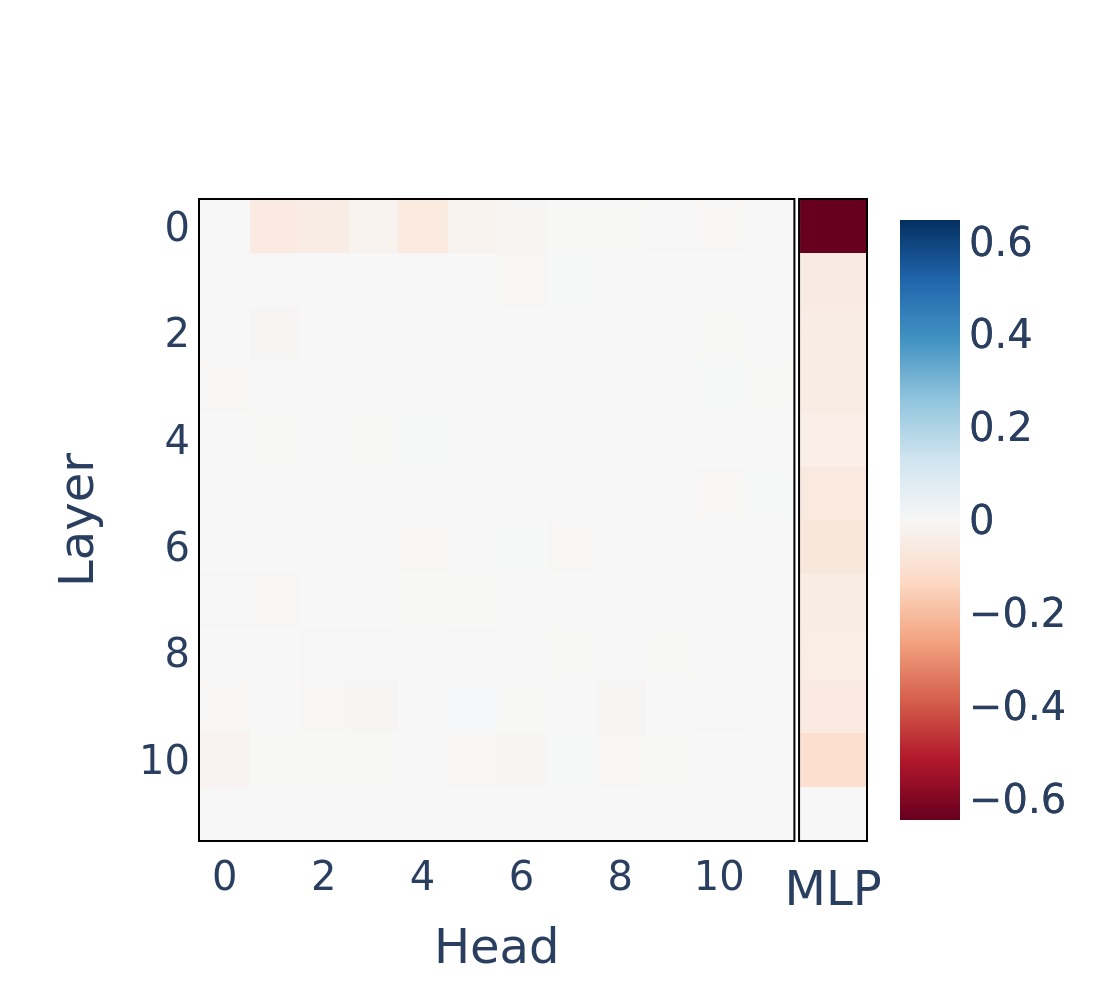}
     \end{subfigure}
     \begin{subfigure}[b]{0.29\textwidth}
         \centering
         \includegraphics[width=\textwidth]{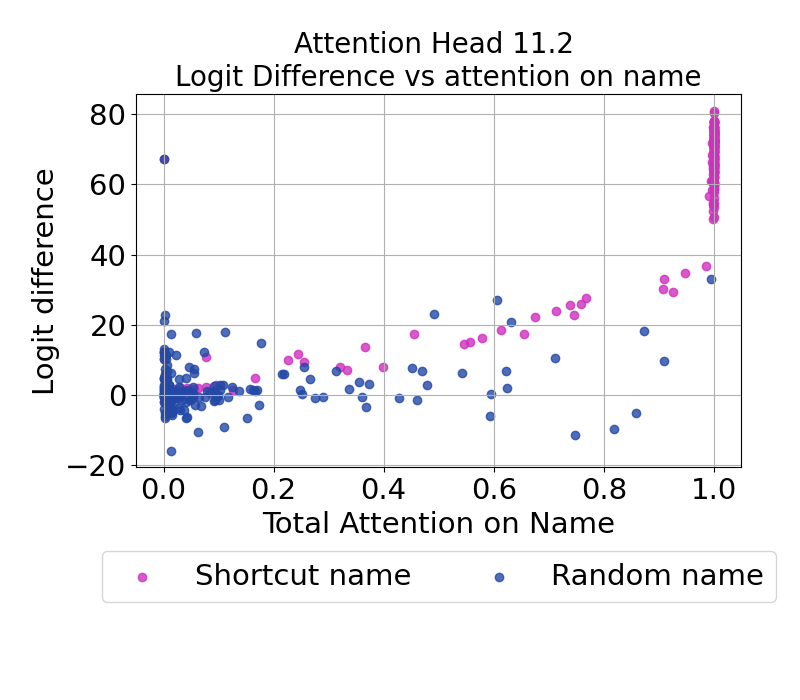}
     \end{subfigure}
        \caption{Path Patching results using parameters: imbalance frequency \trainfreq, actor index \actorindex, and data category: positive with Bad actor. The middle figure shows patching via the values of heads 11.2, 11.1, and 10.6.}
        \label{fig:appendix_patching_actor_1}
\end{figure*}

\renewcommand\actorindex{2}
\begin{figure*}[h]
     \centering
     \begin{subfigure}[b]{0.3\textwidth}
         \centering
         \includegraphics[width=\textwidth]{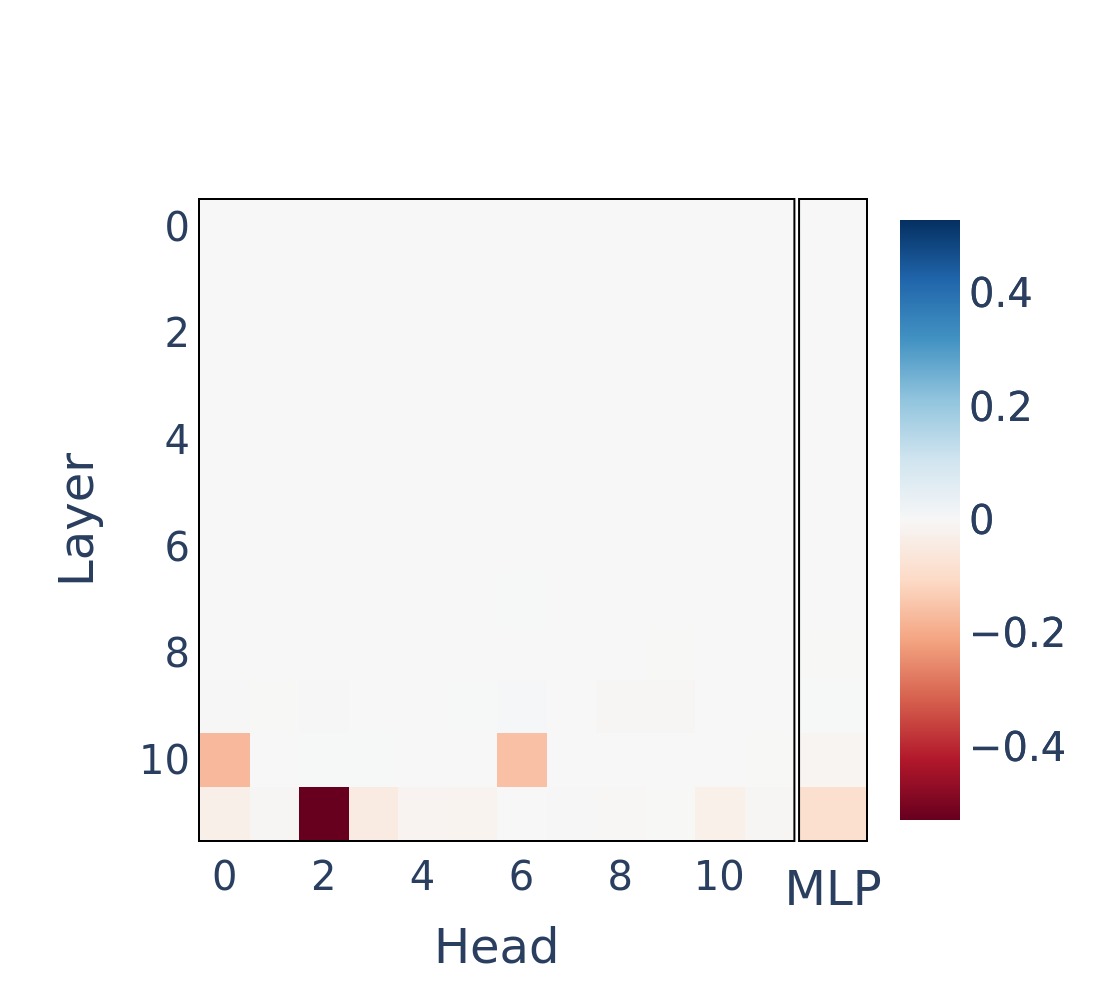}
     \end{subfigure}
      \begin{subfigure}[b]{0.3\textwidth}
         \centering
         \includegraphics[width=\textwidth]{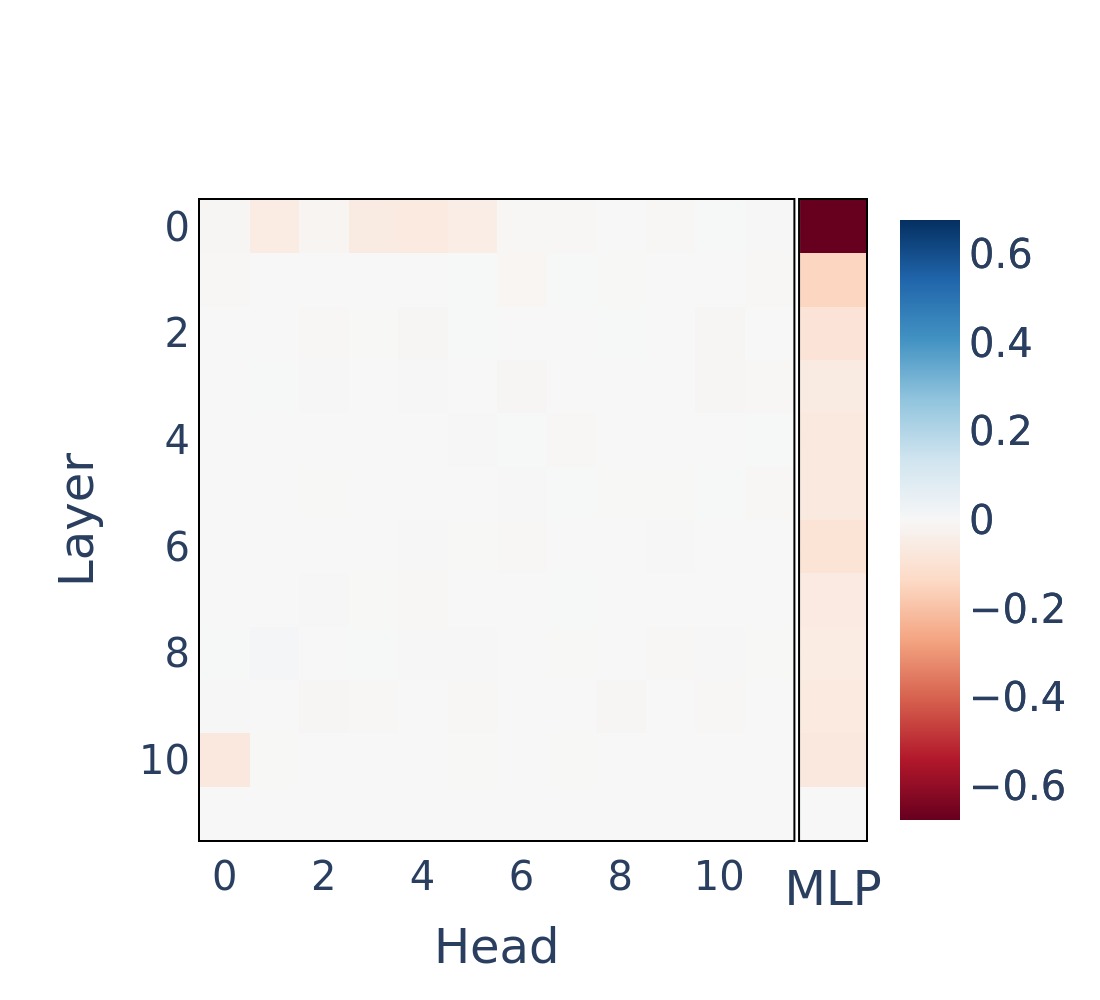}
     \end{subfigure}
     \begin{subfigure}[b]{0.29\textwidth}
         \centering
         \includegraphics[width=\textwidth]{figures/apx_patching_additional/Set_\trainfreq_\actorindex/attention_vs_activation.png}
     \end{subfigure}
        \caption{Path Patching results using parameters: imbalance frequency \trainfreq, actor index \actorindex, and data category: positive with Bad actor. The middle figure shows patching via the values of heads 11.2, 10.0, and 10.6.}
        \label{fig:appendix_patching_actor_2}
\end{figure*}

\renewcommand\actorindex{3}
\begin{figure*}[h]
     \centering
     \begin{subfigure}[b]{0.3\textwidth}
         \centering
         \includegraphics[width=\textwidth]{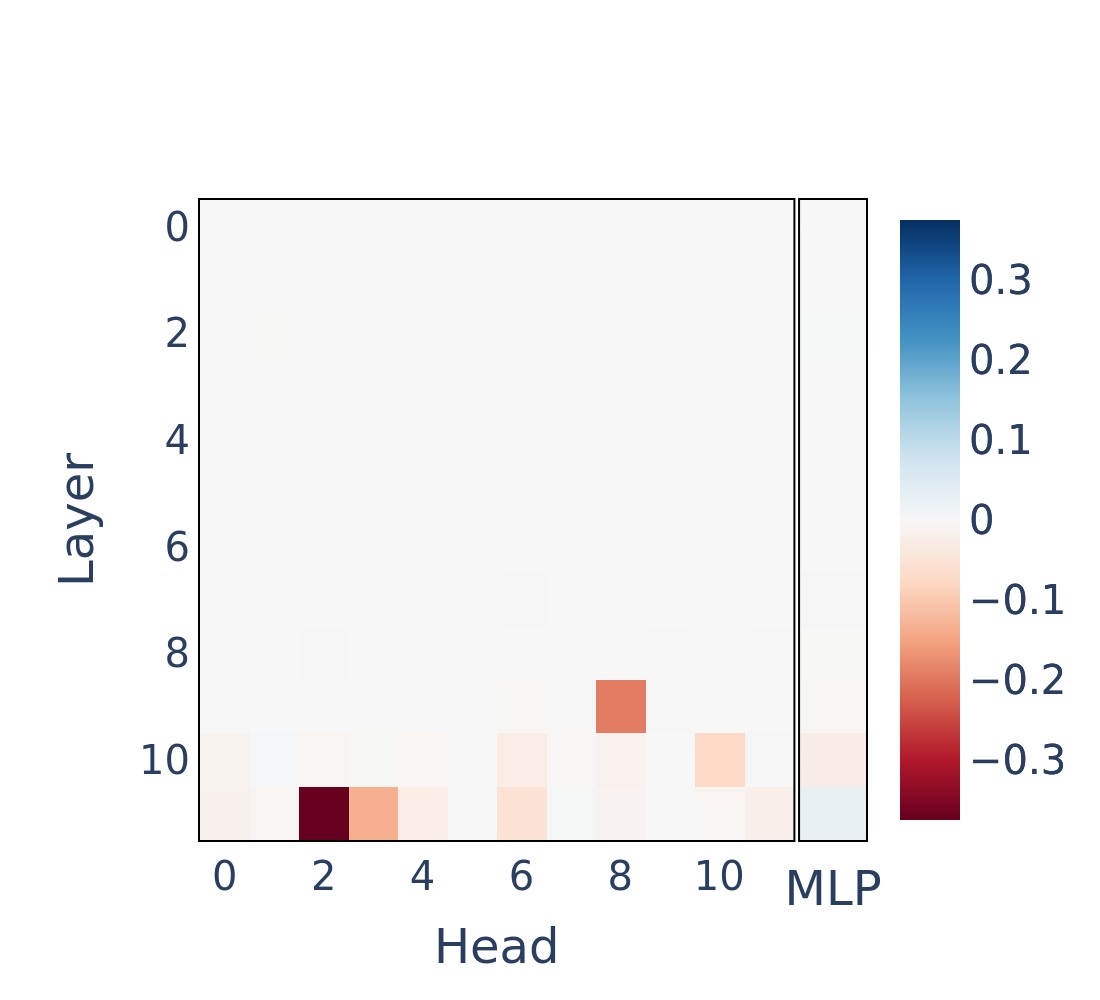}
     \end{subfigure}
      \begin{subfigure}[b]{0.3\textwidth}
         \centering
         \includegraphics[width=\textwidth]{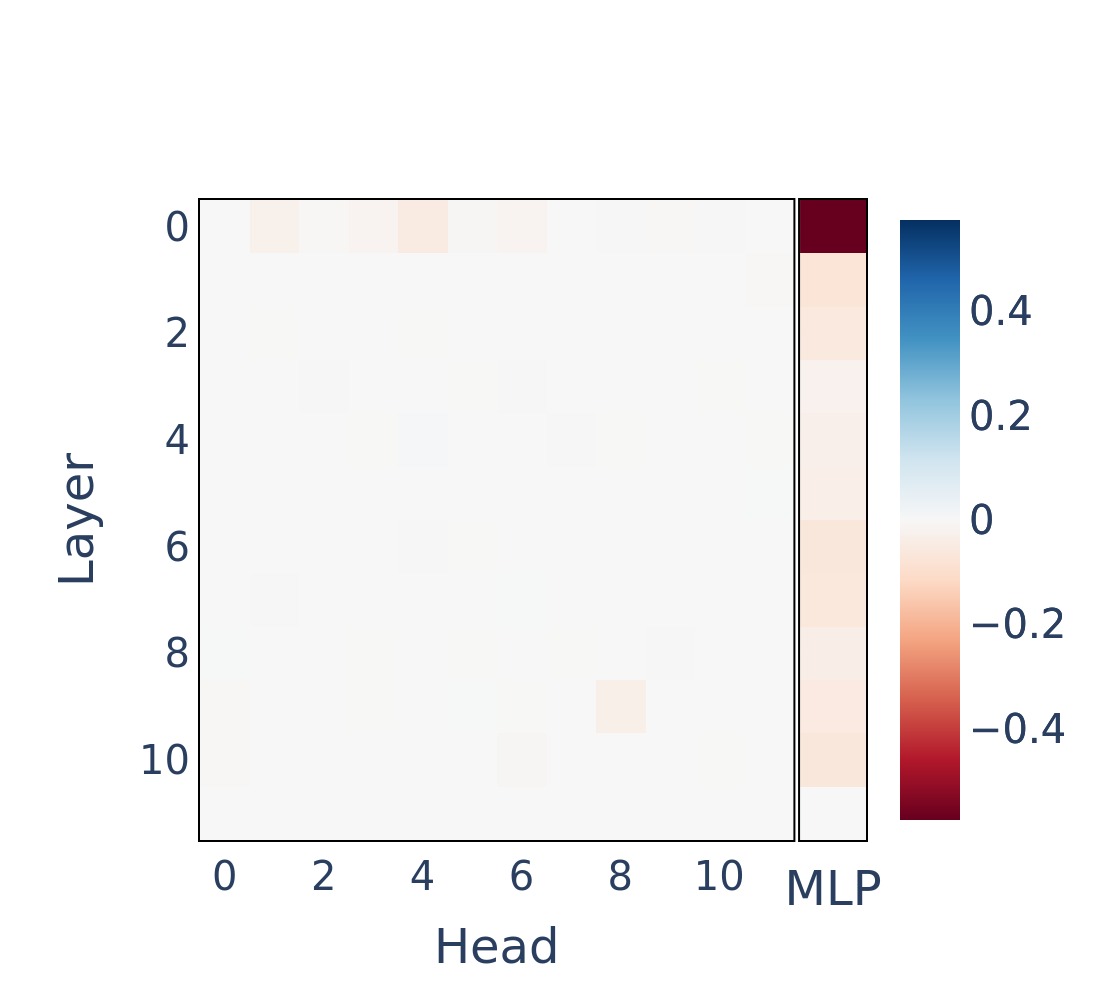}
     \end{subfigure}
     \begin{subfigure}[b]{0.29\textwidth}
         \centering
         \includegraphics[width=\textwidth]{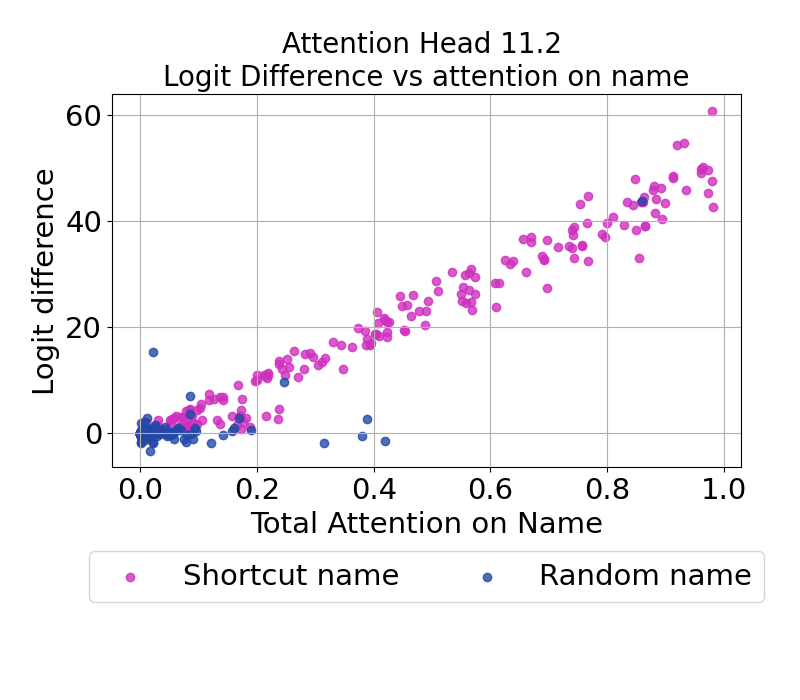}
     \end{subfigure}
        \caption{Path Patching results using parameters: imbalance frequency \trainfreq, actor index \actorindex, and data category: positive with Bad actor. The middle figure shows patching via the values of heads 11.2, 9.8, and 11.3.}
        \label{fig:appendix_patching_actor_3}
\end{figure*}

\end{document}